\documentclass{article}

\usepackage{microtype}
\usepackage{graphicx}
\usepackage{subfigure}
\usepackage{booktabs} % for professional tables

\usepackage{enumitem}
\usepackage{paralist}
\usepackage{comment}
\usepackage{multirow}
\usepackage{diagbox}
\usepackage{amsmath}
\usepackage{url}

\usepackage[accepted]{mlsys2024}

\mlsystitlerunning{HeteroSwitch: Characterizing and Taming System-Induced Data Heterogeneity in Federated Learning}

\begin{document}

\twocolumn[
\mlsystitle{HeteroSwitch: Characterizing and Taming System-Induced Data Heterogeneity in Federated Learning}

% unless you've provided the [accepted] option to the mlsys2024


% \mlsyssetsymbol{equal}{*}

\begin{mlsysauthorlist}
\mlsysauthor{Gyudong Kim}{ko}
\mlsysauthor{Mehdi Ghasemi}{asu}
\mlsysauthor{Soroush Heidari}{asu}
\mlsysauthor{Seungryong Kim}{ko} \\
\mlsysauthor{Young Geun Kim}{ko}
\mlsysauthor{Sarma Vrudhula}{asu}
\mlsysauthor{Carole-Jean Wu}{meta}
\end{mlsysauthorlist}

\mlsysaffiliation{ko}{Korea University.}
\mlsysaffiliation{asu}{Arizona State University.}
\mlsysaffiliation{meta}{Meta (work done while at ASU)}

\mlsyscorrespondingauthor{Young Geun Kim}{younggeun\_kim@korea.ac.kr}

\mlsyskeywords{Federated learning, Data heterogeneity, System heterogeneity, Distributed learning, Fairness, Domain generalization}

\vskip 0.3in

\begin{abstract}
Federated Learning (FL) is a practical approach to train deep learning models collaboratively across user-end devices, protecting user privacy by retaining raw data on-device. In FL, participating user-end devices are highly fragmented in terms of hardware and software configurations. Such fragmentation introduces a new type of data heterogeneity in FL, namely \textit{system-induced data heterogeneity}, as each device generates distinct data depending on its hardware and software configurations. In this paper, we first characterize the impact of system-induced data heterogeneity on FL model performance. We collect a dataset using heterogeneous devices with variations across vendors and performance tiers. By using this dataset, we demonstrate that \textit{system-induced data heterogeneity} negatively impacts accuracy, and deteriorates fairness and domain generalization problems in FL. To address these challenges, we propose HeteroSwitch, which adaptively adopts generalization techniques (i.e., ISP transformation and SWAD) depending on the level of bias caused by varying HW and SW configurations. In our evaluation with a realistic FL dataset (FLAIR), HeteroSwitch reduces the variance of averaged precision by 6.3\% across device types.
\end{abstract}
]



\printAffiliationsAndNotice{}  % leave blank if no need to mention equal contribution
%\printAffiliationsAndNotice{\mlsysEqualContribution} % otherwise use the standard text.

\section{Introduction}
Federated learning (FL) enables mobile devices to collaboratively train a shared machine learning (ML) model while keeping all the raw data on device~\cite{mcmahan2017communication, kairouz2021advances}. As only the model gradients are shared with the cloud servers for updating the shared global model, FL has been considered as a practical way to prevent the privacy leakage~\cite{kairouz2021advances, huba2022papaya}. Although FL has gained much attention in various applications, such as computer vision~\cite{li2022integrated}, voice recognition~\cite{guliani2021training}, health monitoring~\cite{brisimi2018federated}, and recommender system~\cite{hejazinia2022fel}, the following key challenges make FL deployment less practical: high degree of system and data heterogeneity causing unstable and degraded performance of the global model~\cite{ kim2021autofl, li2020federated}. 

\textbf{Data heterogeneity:} In FL, the size and distributions of training data can be significantly heterogeneous depending on the geographical locations, cultural backgrounds, personal habits, and device usage patterns of participating users~\cite{kairouz2021advances}. For example, in a handwriting recognition scenario, users who write the same words exhibit differences in stroke width and slant. Such data heterogeneity can lead to imbalanced and biased model updates~\cite{wu2022node}, eventually degrading the overall model performance~\cite{zhao2018federated} or resulting in disparate accuracy across devices~\cite{li2019fair}. Thus, many prior works have tried to mitigate the data heterogeneity by adopting a regularization method to local models~\cite{li2020federated}, sharing a small amount of public data across local clients~\cite{zhao2018federated, mansour2020three}, or employing weighted averaging for model aggregation~\cite{li2019fair}.

\begin{figure}
  \centering
  \includegraphics[width=0.95\columnwidth]{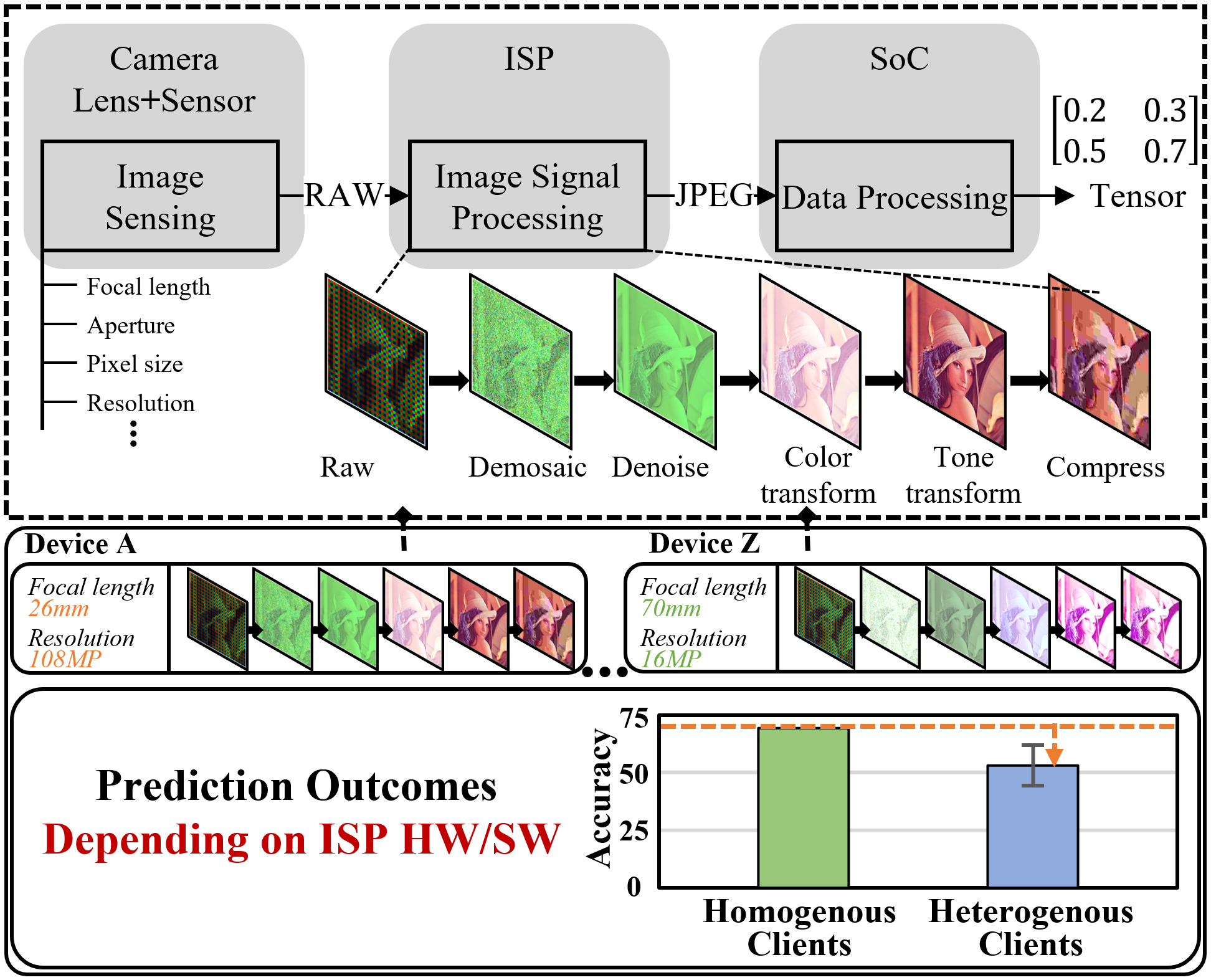}
  \caption{End-to-end ISP pipeline for a vision DNN. Devices generate \textit{heterogeneous data} due to the \textit{SW and HW variations}, degrading model performance significantly.The accuracy of Homogeneous Client is obtained when FL devices are the same, whereas Heterogeneous Client is obtained with different device types.}
  \label{fig:fig1}
\end{figure}

\textbf{System-induced data heterogeneity:} In FL, the hardware of client devices can also be highly diverse: there are more than ten thousand devices with different SoCs, sensors, and software systems in the market~\cite{wu2019machine,kim2020autoscale}. Such system heterogeneity introduces a new type of data heterogeneity, namely \textit{system-induced data heterogeneity}, in FL. For example, in case of vision tasks, heterogeneous sensor hardware can produce noticeably different training data (i.e., images) even for the same scene or object, due to the respective attributes such as focal length, aperture, and other variables (see Figure~\ref{fig:fig1}). The difference across the training data can get even more severe due to the fragmented image signal processing (ISP) algorithms employed onto the devices. This new type of data heterogeneity eventually contributes to bias across the local models, resulting in significant model performance degradation (by 23.5\%, on average, as shown in Fig.~\ref{fig:fig1}). 

Despite the significant impact, system-induced data heterogeneity has been overlooked in FL research. Most prior works have primarily focused on label distribution skew, where a non-IID dataset is formed by partitioning a flat existing dataset based on the labels~\cite{li2020federated,zhao2018federated, mansour2020three,li2019fair,mcmahan2017communication, karimireddy2020scaffold}. Hence, they cannot consider the characteristics of cross-client heterogeneity which should be inherently included in the real-world FL datasets significantly degrading the model performance. 

Furthermore, such system-induced data heterogeneity deteriorates two notable problems in FL: fairness and domain generalization. In FL, fairness involves ensuring that the global-model fairly represents the learning from all devices and is not biased towards any particular group of devices~\cite{pessach2022review, maeng2022towards}. Domain generalization (DG), meanwhile, involves designing machine learning models that can generalize well to new, unseen domains~\cite{wang2022generalizing}. In the context of FL, the domain can be a particular type of devices, each displaying unique characteristics due to system-induced data heterogeneity.

In this paper, we first characterize the impact of system-induced data heterogeneity for FL. We collect a vision dataset using a collection of devices of heterogeneous hardware\footnote{The dataset is available at \url{https://github.com/CASL-KU/HeteroSwitch}.}. By using the collected dataset, we analyze the variations and inconsistencies in the global model that arise due to system-induced data heterogeneity originated from hardware (i.e., sensors) and software  (i.e., ISP algorithms). We also investigate fairness and domain generalization implications which can stem from system-induced data heterogeneity in realistic FL environments. Based on the characterization results, we propose HeteroSwitch --- a selective generalization method that combines the ISP transformation and SWAD --- which counteracts the effects of system-induced data heterogeneity in FL. Compared with the baseline, HeteroSwitch reduces the variance of accuracy across device types by 79.5\%, while improving the worst out-of-distribution (OOD) accuracy by 5.8\%.

\textbf{Key Contributions:}
We summarize the main contributions of this work as follows:
\begin{compactitem}
\item We create a new dataset that takes into account client devices of heterogeneous hardware to independently identify the impact of system-induced data heterogeneity. We attempt to discern the most influential factors contributing to system-induced data heterogeneity in FL models during the data generation process, and examine the impact of \textit{system-induced heterogeneity} on the FL model.
\item We investigate the fairness and domain generalization implications which can be deteriorated by system-induced data heterogeneity.
\item Based on the analysis, we propose HeteroSwitch. By switching the use of generalization techniques in an FL environment where each device possesses heterogeneous data, HeteroSwitch effectively mitigates the model performance degradation caused by system-induced data heterogeneity.
\end{compactitem}
\section{Background}

\subsection{Federated Learning}
Federated Learning (FL) provides a privacy-preserving alternative to conventional machine learning, allowing for collaborative model training across various devices while keeping the data locally stored~\cite{mcmahan2017communication}. Given N local client devices, a server first initializes a global deep learning model and its global parameters by specifying the number of local training epochs E, the local training minibatch size B, and the number of participant devices K. (B, E, K) is determined by FL-based services~\cite{mcmahan2017communication}. In each training round, the server selects \textit{K} devices from the total \textit{N} devices. The global model is then broadcast to the selected devices. Each selected device locally trains the model using its data samples with the batch size \textit{B} over \textit{E} epochs. Once training is completed, the devices send the model gradients back to the server. The server averages these gradients to update the global model. This process is repeated until the desired accuracy is achieved.

\subsection{ISP Pipeline}
In FL, images collected by each client are used for local-model training. Each device produces different images depending on a wide range of hardware and software. Fig.~\ref{fig:fig1} depicts the process from image capturing to the formation of a tensor to be used by the training of a DNN or for inference. (1) At the beginning, the image sensor records the light signal as RAW data based on its properties like focal length, aperture, pixel size, and resolution. (2) After that, a series of image signal processing (ISP) stages (i.e., from Demosaic to Compress in Fig.~\ref{fig:fig1}) are applied to the RAW data to produce a human visible image~\cite{buckler2017reconfiguring, hansen2021isp4ml}.
The ISP stages include:
\begin{compactitem}
\item \textbf{Demosaicing} converts RAW data into a color image.
\item \textbf{Denoising} removes noise from the image.
\item \textbf{Color transformation} (i.e., White Balance adjustments) corrects the colors in the image to make them natural~\cite{wyszecki2000color}.
\item \textbf{Color transformation} (Gamut mapping) converts the colors of the image to a standard gamut.
\item \textbf{Tone transformation} adjusts the brightness and contrast of the image.
\item \textbf{Image compression} reduces file size while attempting to maintain image quality.
\end{compactitem}
(3) Finally, the images are transformed to a tensor to be used by model training or inference.

In FL, the heterogeneous combination of sensors and ISP algorithms from the large collection of client devices cause system-induced data heterogeneity. 
\section{System-induced Data Heterogeneity in FL}
\label{sec:characterize}
This section characterizes the impact of system-induced data heterogeneity on FL. We create a novel image dataset by collecting images using a collection of representative devices of different system hardware characteristics (Section~\ref{sec:datasetcreation}). By using the dataset, we analyze how system-induced data heterogeneity introduces bias on the global model in FL (Section~\ref{sec:breakdown}). We also attempt to discern the most influential factors contributing to system-induced data heterogeneity during the data generation process (Section~\ref{sec:breakdown1} and~\ref{sec:breakdown2}).

\begin{table}[t]
\caption{Mobile devices used in dataset creation, categorized by performance tiers and vendors. (N\%) indicates the market share of respective device. H, M, and L represents high-end, mid-end, and low-end devices, respectively.}
\label{table:devices}
\vskip 0.1in
\begin{center}
  \fontsize{9pt}{10.5pt}\selectfont
  {
\begin{tabular}{|c|ccc|}
\hline
\multirow{2}{*}{Level} & \multicolumn{3}{c|}{Vendor} \\ \cline{2-4} 
 & Samsung & LG & Google \\ \hline
H & GalaxyS22 (12\%) & VELVET (2\%) & Pixel5 (1\%) \\
M & GalaxyS9 (27\%) & G7 (5\%) & Pixel2 (3\%) \\
L & GalaxyS6 (38\%) & G4 (8\%) & Nexus5X (4\%) \\ \hline
\end{tabular}
    }
\end{center}
\vskip -0.1in
\end{table}

\subsection{Dataset Creation}
\label{sec:datasetcreation}
To isolate and examine the impact of system-induced data heterogeneity on FL, we create a custom dataset by using a total of nine smartphones (Table~\ref{table:devices}). We select three smartphones from each of the three different vendors --- Samsung, LG, and Google, representing high-end (H), mid-end (M), and low-end (L) categories with at least two-year gaps between their release dates~\cite{kim2020autoscale, kim2021autofl, wu2019machine} --- by introducing variations across vendors and performance tiers, we emulate a realistic device composition for FL and assess how device heterogeneity contributes to system-induced data heterogeneity.

Each device is fixed with tripod, and used to capture images displayed on a monitor in a dark room --- to isolate the impact of system-induced data heterogeneity and to prevent a potential introduction of the other types of feature distribution skew, we controlled other external factors (e.g., lighting, position, and object being photographed) that may affect the captured images~\cite{cidon2021characterizing, zhang2016image}. For the images, we use 12 non-overlapping ImageNet~\cite{deng2009imagenet} classes from 12 higher-level categories for the images displayed: Chihuahua, Altar, Cock, Abaya, Ambulance, Loggerhead, Timber Wolf, Tiger Beetle, Accordion, French Loaf, Barber Chair, and Orangutan. 

This dataset is compact, yet sufficiently challenging to learn with a higher number of labels and larger image sizes compared to datasets used in previous FL works~\cite{lecun1998mnist, krizhevsky2009learning, cohen2017emnist, scheuerman2021datasets}. To separate the impact of hardware (HW) and software (SW) differences, we collected both RAW data (i.e. unprocessed and uncompressed data directly obtained from the image sensor without employing the ISP algorithms) and images processed by the default camera application of each device.

\begin{table*}[tb]
 \caption{Model quality degradation in model when deployed to various device types, compared to the training device type.}
 \label{table:Data heterogeneity across Device type}
\vskip 0.15in
\centering
\fontsize{9pt}{11pt}\selectfont{
\begin{tabular}{|l|ccccccccc|c|}
\hline
\multicolumn{1}{|c|}{\multirow{2}{*}{\textbf{Train on}}} & \multicolumn{9}{c|}{\textbf{Test on (Model Quality Degradation)}} & \multirow{2}{*}{\textbf{\begin{tabular}[c]{@{}c@{}}Mean\\ Others\end{tabular}}} \\ \cline{2-10}
\multicolumn{1}{|c|}{} & Pixel5 & Pixel2 & Nexus5X & VELVET & G7 & G4 & S22 & S9 & S6 &  \\ \hline
Pixel5 & \textbf{-} & 5.7\% & 21.3\% & 10.0\% & 20.0\% & 21.2\% & 22.4\% & 12.9\% & 24.6\% & 17.3\% \\
Pixel2 & 1.0\% & \textbf{-} & 11.4\% & 5.2\% & 11.8\% & 14.1\% & 19.0\% & 10.5\% & 15.3\% & 11.0\% \\
Nexus5X & 28.1\% & 19.6\% & \textbf{-} & 24.9\% & 6.7\% & 33.9\% & 43.0\% & 20.2\% & 19.8\% & 24.5\% \\
VELVET & 9.4\% & 10.8\% & 15.6\% & \textbf{-} & 11.0\% & 11.9\% & 20.6\% & 4.3\% & 18.7\% & 12.8\% \\
G7 & 32.7\% & 21.3\% & 12.5\% & 17.7\% & \textbf{-} & 19.1\% & 42.7\% & 11.6\% & 17.9\% & 21.9\% \\
G4 & 14.0\% & 17.3\% & 16.7\% & 13.5\% & 15.8\% & \textbf{-} & 26.8\% & 12.0\% & 13.6\% & 16.2\% \\
S22 & 25.8\% & 21.4\% & 25.9\% & 18.4\% & 20.5\% & 26.4\% & \textbf{-} & 21.9\% & 35.8\% & 24.5\% \\
S9 & 29.6\% & 25.5\% & 14.7\% & 10.6\% & 11.1\% & 29.4\% & 50.7\% & \textbf{-} & 16.4\% & 23.5\% \\
S6 & 29.2\% & 24.9\% & 15.5\% & 24.7\% & 9.8\% & 17.2\% & 43.7\% & 17.2\% & \textbf{-} & 22.8\% \\ \hline
\textbf{Mean Others} & 21.2\% & 18.3\% & 16.7\% & 15.6\% & 13.3\% & 21.7\% & 33.6\% & 13.8\% & 20.3\% & 19.4\% \\ \hline
\end{tabular}
}
\vskip -0.1in
\end{table*}

\subsection{Data Heterogeneity across Device type}
\label{sec:breakdown}

In FL, system-induced data heterogeneity can create significant bias across different device types. Table~\ref{table:Data heterogeneity across Device type} shows quality degradation of the global model across various device types, compared to when the model is tested on the same device type it was trained on. The quality of the model is measured by its accuracy on each deployed device. Each row represents the device type used by the clients for training the global model, while each column shows the device type used for testing the trained global model. The Mean Others of each device refers to the average model quality degradation on all devices except the device itself --- the degradation indicates how easily the model is affected by \textit{system-induced heterogeneity} alone. The highest accuracy is always achieved when the model is tested on the same device type it was trained on, and there is an observable drop (1.0\% $\sim$ 50.7\%) in accuracy when the model is tested on the others. 

In Table~\ref{table:Data heterogeneity across Device type}, the model quality degradation depends on the device type. For example, when the model trained on a G7 is tested on Pixel5, the accuracy degrades by 32.7\% --- the accuracy degrades by 20.0\% for the reverse case. This is because the two devices are using significantly different hardware (i.e., camera sensors) and software (i.e., ISP algorithms). On the other hand, Pixel 5 and Pixel 2, which have the smallest differences in HW and SW, show the least model quality degradation when tested on each other (5.7\% and 1.0\%, respectively). %These two devices showed least difference in their Lens\&Sensor specifications among devices and used the same default Camera application (with only minor version differences).
These results demonstrate the unique feature of the \textit{system-induced heterogeneity} which depends on the heterogeneous sensor hardware and ISP algorithms.

To better understand the system-induced data heterogeneity effect, we further investigate its two primary sources in the next subsection: 1) HW (i.e., lens and sensor) variations and 2) SW (i.e., ISP algorithm) variations.

\subsection{Deeper Look at Heterogeneity: HW}
\begin{figure}[t]
  \centering
  \includegraphics[width=0.95\columnwidth]{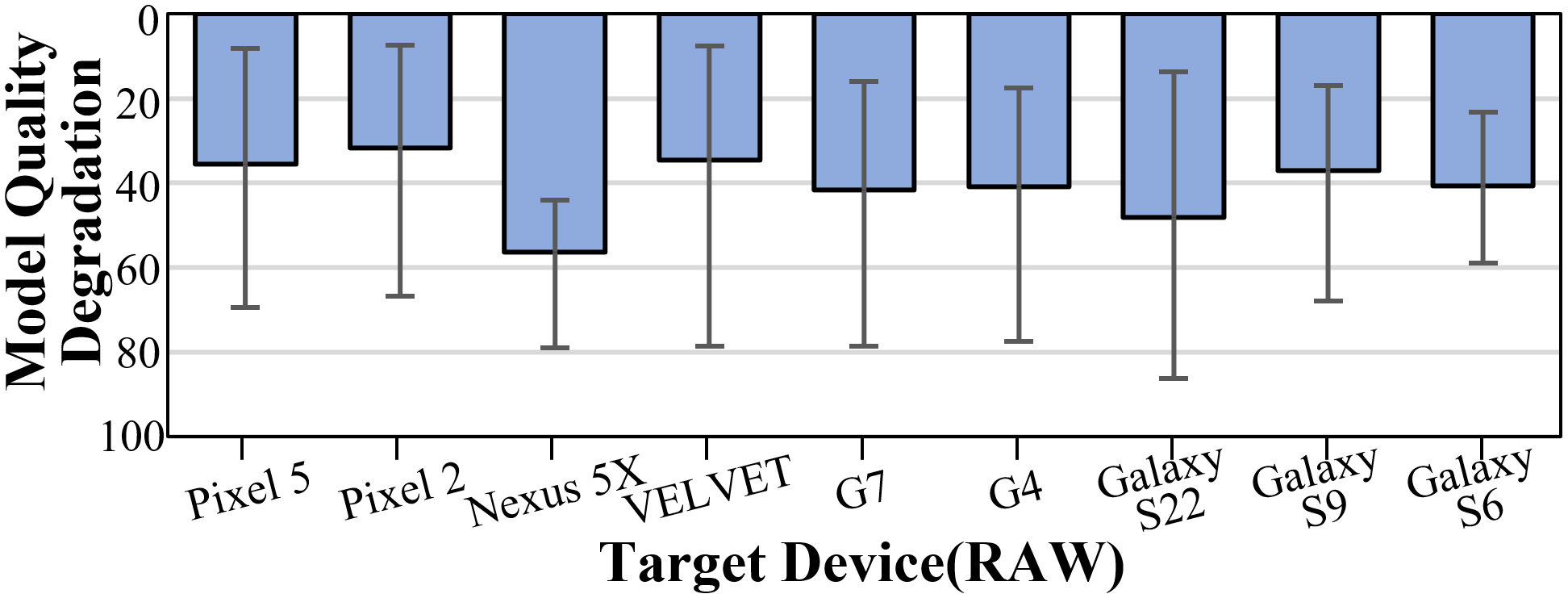}
  \caption{Model quality degradation in model when deployed to various device types using RAW data.}
  \label{fig:breakdown1}
\end{figure}
\label{sec:breakdown1}
In this subsection, we further investigate the impact of HW variations on the data and the trained model performance. To focus on the impact of HW, especially the type of sensors used, we exclude the impact of SW by training the model with RAW data (i.e. unprocessed and uncompressed data directly obtained from the image sensor without employing the ISP algorithms).

The image sensor heterogeneity is a significant source of the data discrepancy. Fig.~\ref{fig:breakdown1} represents the model quality degradation on each target device, when the model is trained with RAW data of the other devices. The x-axis represents the target device, while the y-axis represents the model quality degradation and the error bars indicate the minimum and maximum deviations across the trained devices. Compared to the results obtained using post-processed images (i.e., the last row in Table~\ref{table:Data heterogeneity across Device type}), the degradation is more significant, averaging between 31.74\% and 56.41\%, when we use RAW data. This result implies that this severe heterogeneity among RAW data files should be carefully handled in FL. %The Maximum drop in accuracy exceeds half of its original value in every target file(60.5\%to78.2\%), indicating a severe heterogeneity among RAW data files.

\subsection{Deeper Look at Heterogeneity: SW}
\begin{table*}[tb]
 \caption{ISP algorithms applied to each stage. '\textbf{-}' means we omit this stage.}
 \label{table:ISP algorithms}
\vskip 0.15in
 \centering
 \fontsize{9pt}{9pt}\selectfont
    {
\begin{tabular}{|l|ccc|}
\hline
\multicolumn{1}{|c|}{\textbf{ISP stage}} & \textbf{Baseline} & \textbf{Option 1} & \textbf{Option 2} \\ \hline
Denoising & \begin{tabular}[c]{@{}c@{}}FBDD\\ ~\cite{FBDD}\end{tabular} & \textbf{-} & \begin{tabular}[c]{@{}c@{}}Wavelet-bayesShrink\\ ~\cite{chipman1997adaptive}\end{tabular} \\
Demosaicing & \begin{tabular}[c]{@{}c@{}}PPG\\ ~\cite{lin2003pixel}\end{tabular} & \begin{tabular}[c]{@{}c@{}}Pixel binning\\ ~\cite{binning}\end{tabular} & \begin{tabular}[c]{@{}c@{}}AHD\\ ~\cite{hirakawa2005adaptive}\end{tabular} \\
Color transformation & \begin{tabular}[c]{@{}c@{}}Gray world\\ ~\cite{ebner2007color}\end{tabular} & \textbf{-} & \begin{tabular}[c]{@{}c@{}}White patch\\ ~\cite{ebner2007color}\end{tabular} \\
Gamut mapping & srgb & \textbf{-} & Prophoto \\
Tone transformation & \begin{tabular}[c]{@{}c@{}}srgb gamma correction\\ ~\cite{stokes1996standard}\end{tabular} & \textbf{-} & \begin{tabular}[c]{@{}c@{}}srgb gamma correction\\ +tone Equalization\end{tabular} \\
Image compression & \begin{tabular}[c]{@{}c@{}}JPEG (Quality=85)\\ ~\cite{cidon2021characterizing}\end{tabular} & \textbf{-} & \begin{tabular}[c]{@{}c@{}}JPEG(Quality=50)\\ ~\cite{cidon2021characterizing}\end{tabular} \\ \hline
\end{tabular}
    }
\vskip -0.1in
\end{table*}

\label{sec:breakdown2}
ISP algorithms are applied to RAW data to produce human visible images~\cite{hansen2021isp4ml}. To quantify the contributions of the ISP algorithm variations on system-induced data heterogeneity, we divide the ISP process into six primary stages, from demosaicing to compression, creating distinct images at each stage~\cite{buckler2017reconfiguring}. Table~\ref{table:ISP algorithms} lists the algorithms used at each stage. The impact of each stage on model accuracy is assessed by either omitting a specific stage or applying another algorithm for each stage\footnote{In case of demosaicing, which is a prerequisite for subsequent stages, we use two different methods rather than omitting the stage.} --- we train the global model using data processed by the Baseline column in Table~\ref{table:ISP algorithms}, and test the model while adopting either Option 1 or Option 2 for each stage.

\begin{figure}[t]
  \centering
  \includegraphics[width=0.95\columnwidth]{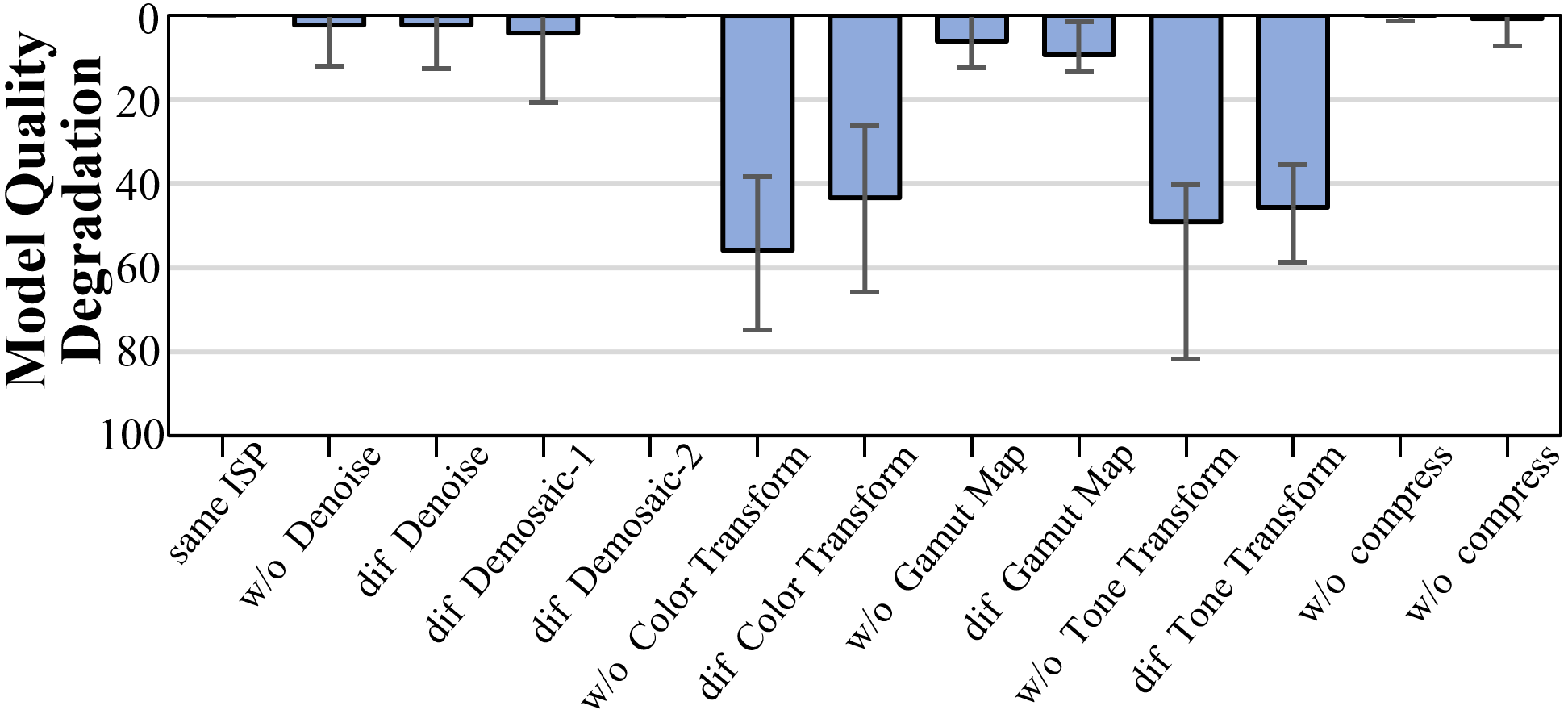}
  \caption{Model quality degradation in model when tested with adjustment of ISP algorithms at every stage.}
  \label{fig:breakdown2}
\end{figure}

Although the ISP algorithms are known to reduce the HW variations~\cite{ebner2007color, morovivc2008color}, variations in certain ISP methods still introduce unmanageable heterogeneity in images that have a detrimental impact on model performance. Fig.~\ref{fig:breakdown2} depicts the model quality degradation due to the various ISP algorithms. The x-axis represents the stage omitted (or modified) from the baseline. As shown in Fig.~\ref{fig:breakdown2}, the model quality degrades when we omit (or modify) each stage of the ISP algorithms. Notably, omitting or modifying the color and tone transformation algorithms result in significant declines in model quality, regardless of the device type --- the accuracy is degraded by 56.0\% and 49.2\% when we exclude Color (specifically, White Balancing) and Tone transformation from ISP stages, respectively, which is even more severe compared to that we observe in RAW data. This result implies that each stage of ISP algorithms, especially the color and tone transformation stages, significantly contributes to system-induced data heterogeneity and thus there is a demand for new solutions to tackle ISP heterogeneity across the devices in FL.
\section{Fairness and Domain Generalization Issues}
\label{sec:ProbleminFL}
In a real-world FL environment, a global model is trained with a number of client devices. Considering our observations in Section~\ref{sec:breakdown}, the diverse types of client devices can contribute differently to the global model due to their unique data characteristics stemming from HW (Section \ref{sec:breakdown1}) and SW variations (Section \ref{sec:breakdown2}). As a result, certain characteristics affected by system heterogeneity can make the global model learn a bias towards those characteristics. This potential issue can be interpreted through the lens of two well-established fields in machine learning: fairness and domain generalization.

\subsection{Fairness}
\begin{figure}[tb]
  \centering
  \includegraphics[width=0.95\columnwidth]{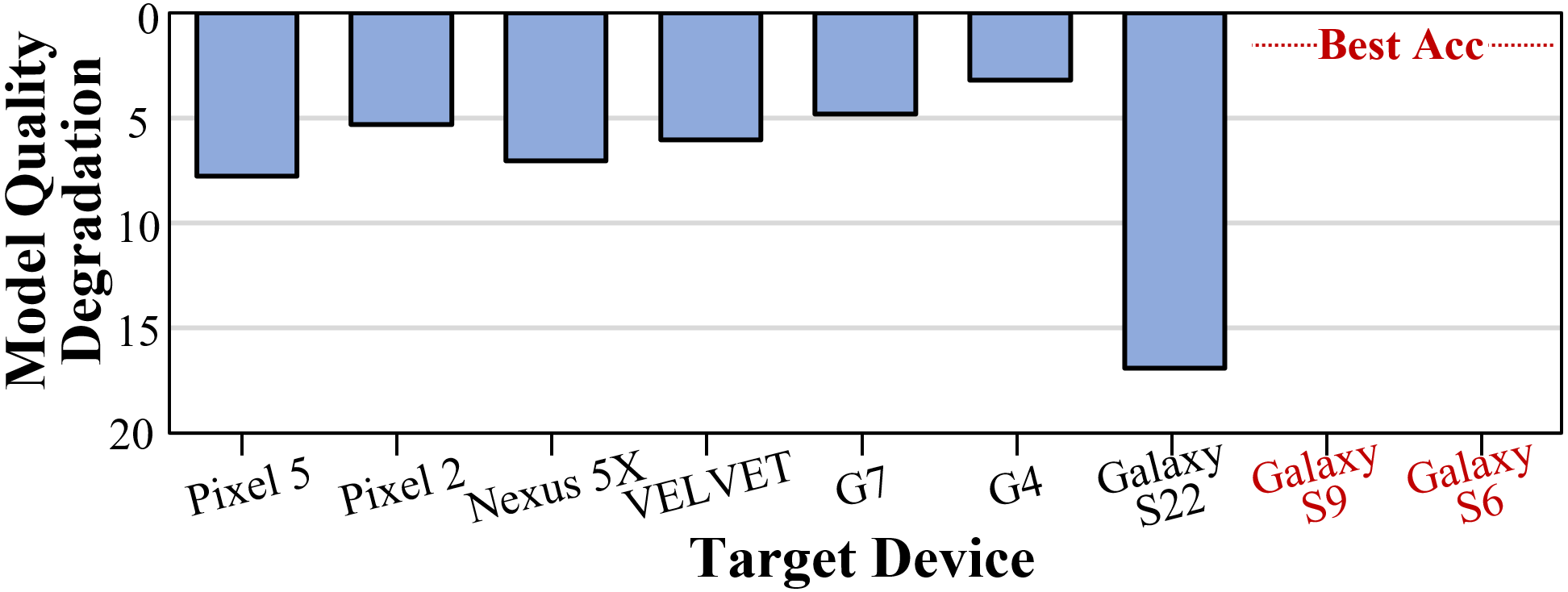}
  \caption{Model quality degradation over the highest accuracy achieved by dominant devices, Galaxy S9 \& S6, showing the bias in the global model towards dominant devices. The participant devices used for training followed the ratio specified in Table~\ref{table:devices}.}
  \label{fig:fair}
\end{figure}

In FL, the number of client devices is not always same for all device types. This uneven distribution makes the fairness problems\footnote{In the realm of FL, fairness ensures the global model performs adequately across all participating devices, rather than being skewed towards a certain type or group of devices.} in FL~\cite{mohri2019agnostic, maeng2022towards}. For example, if a majority of training data comes from a particular group of devices with the similar HW and SW configurations, the global model may become more tailored towards those devices, degrading accuracy for the rest of the devices. To quantify the fairness problem in a realistic manner, we allocate device types of clients according to the vendor market share~\cite{statcounter} (summarized in Table~\ref{table:devices}) and system performance distribution~\cite{wu2019machine}. Note, in our analysis, we refer the device types with the highest percentage of participation as the dominant devices (i.e., Galaxy S9 and S6 in Table~\ref{table:devices}) --- these can be seen as a privileged group in the context of fairness, benefiting from biases in the global model~\cite{pessach2022review}. To assess the fairness of FL across various device types, we compute the model quality degradation on each remaining device compared to the accuracy on the dominant devices.

Fig.~\ref{fig:fair} shows the model quality degradation for each device compared to the dominant devices --- the x-axis represents the device type the model is deployed. As shown in Fig.~\ref{fig:fair}, the accuracy on the 7 less dominant devices is 3.2\% to 16.9\% lower than that on the dominant devices. This implies that the global model has a bias toward the HW specification and SW algorithms of the dominant devices. On the other hand, although Galaxy S22 is the third most used for training, deploying the model to it yields the lowest accuracy among the devices. This also implies that a higher participation rate does not always guarantee a high accuracy, meaning that there exist the system features, such as the advanced ISP algorithms, which can smooth out (or deteriorate) the bias.

\subsection{Domain Generalization}  

In the context of FL, deploying the model to a new, unseen device type is closely related to the concept of domain generalization (DG)~\footnote{Given that more than 500 new smartphones are released every year~\cite{gsmarena}, it is common to see unseen devices for FL-based services.}. This concept involves designing ML models that can effectively generalize to new, unseen domains~\cite{wang2022generalizing}, such as unique device types. In FL, domains can be characterized by system-induced data heterogeneity, where each device exhibits distinct features. To emulate the DG problem in FL, we consider each device as an unseen domain --- note we analyze the ability of the model to generalize across different device types by training the global model with all other devices and subsequently testing it on the selected unseen device.
 
\begin{figure}[t]
  \centering
  \includegraphics[width=0.95\columnwidth]{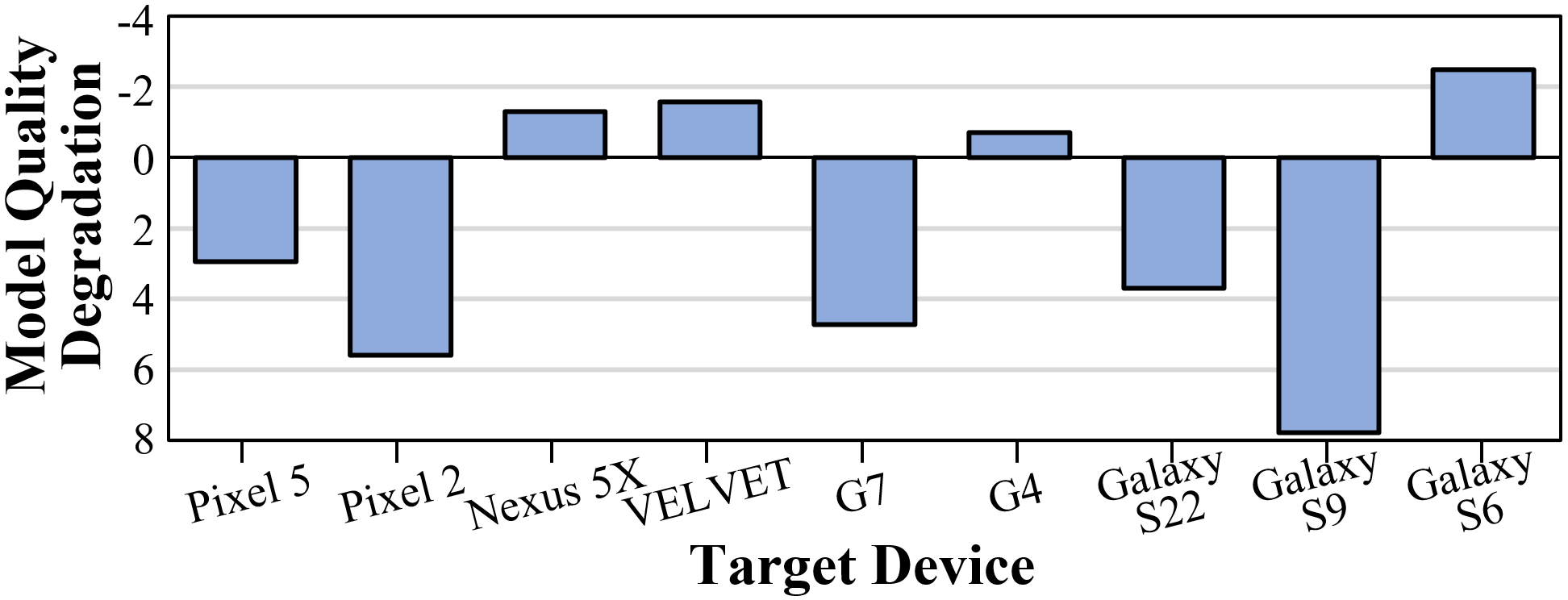}
  \caption{Model quality degradation when a device type is excluded from training, illustrating the complex relationship of DG and accuracy with various device types in FL.}
  \label{fig:DG}
\end{figure}

Fig.~\ref{fig:DG} shows the model quality degradation of global model accuracy when the device is excluded from training, compared to when all device types equally participate in FL (this practical is widely used in DG~\cite{dou2019domain,segu2023batch}). The x-axis represents the device type that was excluded from training, while the y-axis shows the model quality degradation on the excluded (i.e., unseen) device. In DG, it is typically expected that a domain excluded from training would show lower accuracy, since the global model lacks exposure to its specific characteristics~\cite{dou2019domain, segu2023batch}. Interestingly, excluding a device from training does not consistently affect accuracy as they generate different details for samples, such as color, texture, and contextual details~\cite{tommasi2017deeper}. For example, the accuracy for S9 drops when it is not part of the training.

Conversely, older devices like S6, Nexus5X and G4, which have lower resolutions and simpler ISP algorithms, commonly exhibit increased accuracy, even though they were excluded from training. This inconsistent result demonstrates system-induced data heterogeneity can deteriorate the complexity of DG in FL.

Our observations highlight the importance of considering system-induced data heterogeneity and its potential impact on fairness and DG in FL. There is a need for a careful strategy that ensures stable performance of global model for various device types. 

\section{Proposed Design: HeteroSwitch}
\label{sec:Solution}
\vskip 0.1in

\begin{figure*}[ht]
\centering
\includegraphics[width=0.95\textwidth]{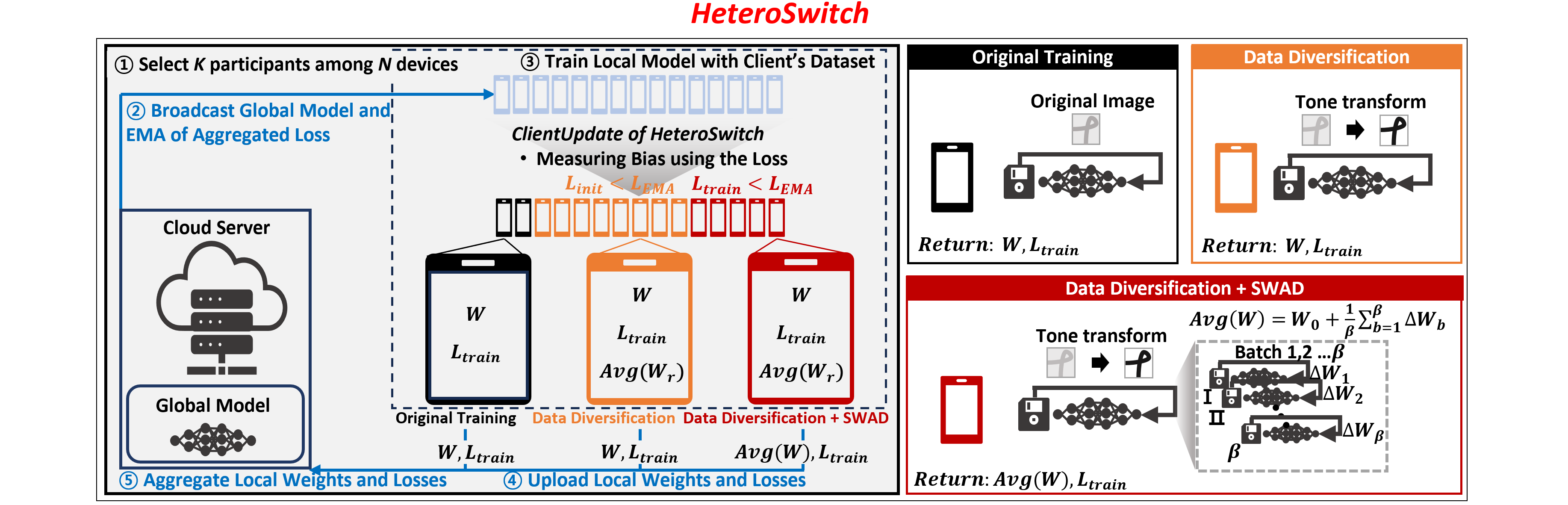} % Reduce the figure size so that it is slightly narrower than the column.
\caption{Overall Design of HeteroSwitch} 
\label{fig:heteroswitch}
\vskip 0.1in
\end{figure*}

%Fig~\ref{fig:heteroswitch} shows the overview of the HeteroSwitch.
As we observe in Section~\ref{sec:ProbleminFL}, system-induced data heterogeneity causes significant accuracy degradation of the deployed DNNs on the typical device types. To mitigate the problem, we propose HeteroSwitch --- a selective generalization technique. Fig.~\ref{fig:heteroswitch} shows the overview of HeteroSwitch. As shown in Fig.~\ref{fig:heteroswitch}, HerteroSwitch incrementally applies generalization techniques to the clients that have biased data due to system-induced data heterogeneity. For each round, HeteroSwitch measures the amount of bias in the data of participating clients, by comparing their initial loss $L_{init}$ with the Exponential Moving Average (EMA) of the aggregated loss $L_{EMA}$.(Section~\ref{sec:solution1}). Based on the measured bias, HeteroSwitch determines whether it applies the generalization techniques to the client data or not (Section~\ref{sec:solution2}). The algorithm detail is explained in Algorithm~\ref{alg:algorithm1}.

\begin{algorithm}[tb]
\caption{ClientUpdate of HeteroSwitch}
\label{alg:algorithm1}
\textbf{Input}: Model Parameters $W$, Client's Dataset $D$, EMA of Aggregated Loss $L_{EMA}$(eq.~\ref{eq:l_bench})\\
\textbf{Parameter}: Learning Rate $\eta$, \# of Epochs $E$, Batch Size $B$\\
\textbf{Output}: Updated Model Parameters $W_{return}$, Train Loss $L_{train}$\\
\textbf{ClientUpdate($W, L_{EMA}$):}
\begin{algorithmic}[1]
    \STATE {$Switch_{1}, Switch_{2} = False$}
    \STATE Calculate $L_{init} = L$($D$, $W$) 
    \IF {$L_{init} < L_{EMA}$}
        \STATE {$Switch_{1} = True$}
    \ENDIF

    \IF {$Switch_{1} == True$}
        \STATE \textit{Random Transformation on $D$(eq.~\ref{eq:random_white_balance}, eq.~\ref{eq:random_gamma})}
    \ENDIF
    \STATE $\beta \leftarrow$ (split $D$ into batches of size $B$) %사실 이부분도 줄일 수 있다. batchsize가 아닌, for batch in ~ 고칠 수 있는 지 확인
    \STATE Initialize $W_{SWA}$ as copy of $W$
    \STATE Initialize batch index $Idx_{b}=0$, Train Loss $L_{train}=0$ %Loss 구하는 부분을 줄일 수 있을 것 같다.
    \FOR {epoch $e=1$ to $E$}
        \FOR {batch $b \in \beta$}
            \STATE $L_{train} \leftarrow \frac{L(b, W)*Idx_{b} + W}{Idx_{b}+1}$
            \STATE $W \leftarrow W - \eta \nabla L(b, W)$
            \IF {$Switch_{1} == True$}
            \STATE $W_{SWA} \leftarrow \frac{W_{SWA}*Idx_{b} + W}{Idx_{b}+1}$
            \ENDIF
            \STATE $Idx_{b} \leftarrow Idx_{b} + 1$ %batch 수정을 통해 줄일 수 있다.
        \ENDFOR
    \ENDFOR
    \IF {$Switch_{1} == True$ \text{ and } $L_{train} < L_{EMA}$ }
        \STATE {$Switch_{2} = True$}
    \ENDIF            
    \IF {$Switch_{2} == True$}
        \STATE $W_{return} = W_{SWA}$
    \ELSE
        \STATE $W_{return} = W$
    \ENDIF
\STATE return $W_{return}$, $L_{train}$ to server 
\end{algorithmic}
\end{algorithm}

\subsection{Bias Measurement}
\label{sec:solution1}
In the context of FL, every client has a unique dataset --- \textit{each data differently contributes to the global model}. Moreover, The dynamic nature of FL causes the datasets used for training not to remain constant every round~\cite{mcmahan2017communication}. With this constraint, applying a "one-size-fits-all" approach may not yield the best results. For example, applying the same level of generalization technique to all clients could negatively affect the learning process for those using less common device types, leading to unnecessary performance degradation for them. To mitigate these effects, HeteroSwitch selectively employs the generalization techniques to the data of participating clients based on the training loss. We use the Exponential Moving Average (EMA) loss from previous communication rounds or the validation loss as the criteria for switching the generalization techniques on the client side.\\
\textbf{EMA of Previous Train Loss $L_{EMA}$: }
\begin{equation}
    \label{eq:l_bench}
    \begin{aligned}
    % & \text{With EMA smoothing factor $\alpha$}, \\
    & L_{EMA,t+1} \leftarrow \alpha * L_{cur} + (1-\alpha) * L_{EMA,t} \\
    \end{aligned}
\end{equation}

If the test loss on dataset of a client before updating the local model is lower than the aggregated training loss calculated during the previous round, the dataset is considered to have a potential bias. This indicates that the characteristics of the dataset are already learned by the global model.

\subsection{Generalization Techniques}
\label{sec:solution2}
Fairness and DG share similar objectives, as discussed in \cite{creager2021environment}. To solve the fairness problem, it is important to elevate the accuracy of devices that underperform. On the other hand, the domain generalization problem seeks to improve the accuracy for unseen devices. In essence, if consistent accuracy is achieved across all devices, regardless of their presence during training, it may be sufficient to address both fairness and DG problems. To achieve this, we propose a two-pronged method focusing on the dataset diversification and the model generalization. The method is used on the client-side during the training, adapting to system-induced data heterogeneity.

\textbf{Dataset Diversification with ISP Adjustment: }
Given heterogeneity across the data from different device types, we propose to expand the diversity of the data via \textit{random transformation} during training --- such transformation, including geometric, color adjustments, and random erasing, does not change the inherent label or meaning of the image, encouraging the model to learn from more varied range of data~\cite{shorten2019survey}. 

Motivated by our observations in Section~\ref{sec:breakdown2}, which highlighted the need for addressing Tone and Color (especially White Balance~\cite{wyszecki2000color}) variations, we temporarily adopt random White Balance and tone transformations (via random gamma application~\cite{wyszecki2000color}) to the dataset of each client during local training.\\
\textbf{Random WB:}
\begin{equation}
    \label{eq:random_white_balance}
    \begin{aligned}
    & \text{With $r_1,r_2,r_3 \sim U(1-degree,1+degree)$}, \\
    & \begin{bmatrix} R\\ G\\ B \end{bmatrix}_{out} = \begin{bmatrix} r_1&0&0\\ 0&r_2&0\\0&0&r_3 \end{bmatrix}\begin{bmatrix} R\\ G\\ B \end{bmatrix}_{in} \\
    \end{aligned}
\end{equation}
\textbf{Random Gamma:}
\begin{equation}
    \label{eq:random_gamma}
    \begin{aligned}
        & \text{With $\gamma \sim U(1-degree,1+degree)$}, \\
        & \text{Img}_{out} = \text{Img}_{in}^{\gamma}\\
    \end{aligned}
\end{equation}

However, data diversification does not necessarily need to be applied to every data. Given distinct ISP methods employed by each device, some data may pass through ISP stages that are not learned by the global model. To make the global model learn diverse characteristics of the ISP stages, we do not employ the data diversification in such a case.

\begin{figure}[t]
\centering
\includegraphics[width=0.95\columnwidth]{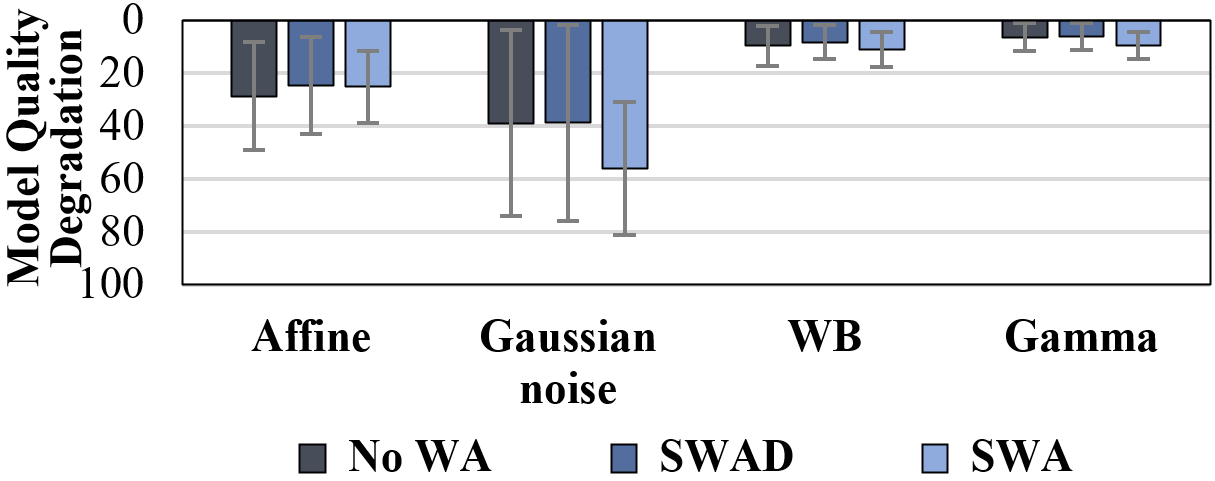}
\caption{Comparison of Model Quality Degradation for different training methods and transformations} 
\label{fig:a1}
\end{figure}

%Ver. 2 SWA를 Dataset Diversification에 종속적이게.
\textbf{Model Generalization through SWAD: }
Even with data diversification that accounts for ISP adjustments, biases may still arise to the global model due to extreme differences between devices. We employ a weight averaging method~\cite{izmailov2018averaging, cha2021swad} for further generalization. Fig.~\ref{fig:a1}\footnote{For this experiment, we use the original 12-class ImageNet dataset that we utilized for dataset creation (Section~\ref{sec:datasetcreation}). In each training scenario, we train the model for 10 epochs, applying a random data transformation at a low degree (degree=0.3). After training, we measure the accuracy on the original dataset. Then, we compare this accuracy with the accuracies on the transformed datasets (degrees ranging from 0.3 to 0.9) to assess the robustness of the model.} shows the comparative robustness of two different weight averaging methods, Stochastic Weight Averaging Densely (SWAD)~\cite{cha2021swad} and conventional SWA~\cite{izmailov2018averaging}, for three training scenarios: applying only data random transformation, applying SWAD with transformation, and applying conventional SWA with transformation. The x-axis shows different random transformation methods.

Utilizing diverse data transformations with SWAD considerably enhances the model robustness. As depicted in Figure~\ref{fig:a1}, SWAD with data transformations consistently exhibits superior robustness across all transformations. This is highlighted by observed improvements in metrics for Affine, Gaussian noise, WB, and Gamma by 14.0\%, 0.4\%, 14.6\%, and 3.26\%, respectively, over models trained only with transformations. The application of SWA also provides better resilience against the Affine transformation, showing a 12.0\% improvement over models using only transformations without weight averaging. However, transformations with SWA shows increased vulnerability to transformations such as Gaussian noise, WB, and Gamma, resulting in greater quality degradation. Consequently, SWAD shows better generalization performance compared to SWA.

SWA averages the model weights per every epoch while SWAD averages the model weights per every batch which is coarser than epoch. Hence, the integration of SWAD with random transformation, designed to average out significant variability with fewer samples~\cite{dekking2005modern}, enables DNN models to have enhanced adaptability to both geometric (e.g., Affine) and appearance-based (e.g., WB \& Gamma) variations during training.

\section{Evaluation Results and Analysis}
\label{sec:evaluation}
To assess the effectiveness of HeteroSwitch, we evaluate its impact on both fairness and domain generalization in FL by using the test accuracy of the global model~\cite{shi2023towards} on each deployed device type. For fairness, we measure the average variance of accuracy across devices --- the portion of participation of each device type follows the market share summarized in Table.~\ref{table:devices}. For DG, we use the worst-case accuracy across devices as the metric to ensure a minimum required performance across all devices~\cite{sagawa2019distributionally, krueger2021out}. 

We configure FL experiments with total devices \( N = 100 \), minibatch size \( B = 10 \), selected devices for training for each round \( K = 20 \), local epoch \( E = 1 \), and number of rounds \( T = 1000 \). For a DNN model, we use MobileNetv3-small~\cite{howard2019searching}, which is widely used in mobile execution environments. We implement HeteroSwitch using PyTorch, and compare it with a baseline method, FedAvg~\cite{mcmahan2017communication}, and prior works for data heterogeneity: q-FedAvg~\cite{li2019fair}, FedProx~\cite{li2020federated}, and Scaffold~\cite{karimireddy2020scaffold}.

We also evaluate HeteroSwitch with a realistic FL dataset, Flair~\cite{song2022flair} --- this dataset includes images collected by real end-users with more than one thousand device types. Since Flair targets multi-label classification, we compare the averaged-precision across device types and variance of HeteroSwtich with a baseline of FedAvg and prior works (i.e., q-FedAvg and FedProx).

\begin{table}[ht]
\caption{Evaluation of HeteroSwitch on fairness and DG}
\label{table:proposed_solutions}
\vskip 0.15in
\begin{center}
\fontsize{9pt}{11pt}\selectfont
{%
\begin{tabular}{|l|c|cc|}
\hline
\multirow{2}{*}{\textbf{Method}} & \textbf{DG} & \multicolumn{2}{c|}{\textbf{Fairness}} \\ \cline{2-4} 
 & \begin{tabular}[c]{@{}c@{}}Acc (\%)\\ (worst case)\end{tabular} & Variance &\begin{tabular}[c]{@{}c@{}}Acc (\%)\\ (average)\end{tabular} \\ \hline
(Baseline) FedAvg  & 61.17 & 8.63 & 64.01 \\ \hline
ISP Transformation & 63.83 & 2.58 & 66.81 \\
+ SWAD & 64.50 & 2.74 & 66.29 \\
HeteroSwitch & \textbf{64.71} & \textbf{1.77} & \textbf{67.38} \\ \hline
q-FedAvg & 59.38 & 5.40 & 64.95 \\
FedProx & 61.50 & 8.43 & 65.16 \\ 
Scaffold & 57.50 & 7.53 & 66.18 \\ \hline
\end{tabular}%
}
\end{center}
\vskip -0.15in
\end{table}

\subsection{HeteroSwitch Evaluation}

HeteroSwitch shows a significant variance and accuracy improvement in both fairness and DG as depicted in Table~\ref{table:proposed_solutions}. Notably, the use of our proposed generalization methods - ISP (WB\&Tone) transformation and SWA per every update - in a incremental manner leads to the improvement compared to the one-fits-all counterparts. As the ISP transformation adds further diversity into the dataset, it reduces the  4.4\% higher worst-case accuracy over the baseline, handling the DG problem with \textit{system-induced heterogeneity}. It also mitigates the difference between learned device types due to ISP, improving variance and average accuracy by 68.2\% and 4.4\%, respectively. per-batch SWA further improves the worst-case accuracy by 1.0\% showing better generalization capability --- it is designed for stronger generalization technique by averaging updates with different ISP transformation. However, this degrades the variance and average accuracy by 6.2\% and 0.8\%, respectively, compared to ISP transformation alone. This is because a one-size-fits-all excessive generalization can cause the unnecessary performance degradation for marginalized devices during training. 

HeteroSwitch overcomes this limitation by selectively applying the generalization techniques based on the loss comparison with the previous training rounds. As a result, it shows the highest worst-case accuracy, i.e., 5.8\% over the baseline, demonstrating the highest generalization capability. HeteroSwitch also demonstrates its ability to handle marginalized devices, showing the best variance and average accuracy (79.5\% and 5.3\% over the baseline, respectively).

\subsection{Comparison with Prior Works}
\label{sec:Compare}
We compare HeteroSwitch with three prominent FL prior works: q-FedAvg~\cite{li2019fair}, FedProx~\cite{li2020federated}, and Scaffold~\cite{karimireddy2020scaffold}. 

\textbf{q-FedAvg} aims to minimize the accuracy variance among clients owing to data heterogeneity, by assigning weights to each client dataset based on the loss. However, q-FedAvg does not consider system-induced data heterogeneity across different device types in FL. As a result, it fails to generalize to unseen device type with 2.9\% decrease in worst-case accuracy; it improves the variance and average accuracy by 37.4\% and 1.5\%, respectively, over the baseline though. 

\textbf{FedProx} mitigates data heterogeneity in FL by adjusting the magnitude of local updates with an additional L2 regularization. Similar to q-FedAvg, this method does not focus on system-induced data heterogeneity. Hence, although it improves the worst-case accuracy, variance and average accuracy by 0.5\%, 2.3\%, and 1.8\%, respectively, over the baseline, it shows lower performance compared to HeteroSwitch.

\textbf{Scaffold}, which employs client control variates, to adjust global update directions,  addresses non-IID data variance. Still, it faces a similar challenge to q-FedAvg in generalizing to unseen devices, with a 6.0\% drop in worst-case accuracy.

\textbf{HeteroSwitch}, which is designed to effectively handle system-induced data heterogeneity, shows better variance, average accuracy, and worst-case accuracy, compared to all the above methods. This distinction highlights the unique contribution of HeteroSwitch in addressing heterogeneous device environments of FL, providing a more balanced and generalized solution.

\subsection{Effectiveness to DNN Models}

\begin{table*}[tb]
\caption{Evaluation of HeteroSwitch with different Model architectures}
\label{table:accuracyofarch}
\vskip 0.15in
\centering
\fontsize{9pt}{10pt}\selectfont
\begin{tabular}{|l|ccc|ccc|}
\hline
\multirow{3}{*}{\textbf{Models}} & \multicolumn{3}{c|}{\textbf{FedAvg}} & \multicolumn{3}{c|}{\textbf{HeteroSwitch}} \\ \cline{2-7} 
 & \multicolumn{1}{c|}{\textbf{DG}} & \multicolumn{2}{c|}{\textbf{Fairness}} & \multicolumn{1}{c|}{\textbf{DG}} & \multicolumn{2}{c|}{\textbf{Fairness}} \\ \cline{2-7} 
 & \multicolumn{1}{c|}{\begin{tabular}[c]{@{}c@{}}Acc (\%)\\ (worst case)\end{tabular}} & Variance & \begin{tabular}[c]{@{}c@{}}Acc (\%)\\ (average)\end{tabular} & \multicolumn{1}{c|}{\begin{tabular}[c]{@{}c@{}}Acc (\%)\\ (worst case)\end{tabular}} & Variance & \begin{tabular}[c]{@{}c@{}}Acc (\%)\\ (average)\end{tabular} \\ \hline
MobileNetV3-small & \multicolumn{1}{c|}{61.17} & 8.63 & 64.01 & \multicolumn{1}{c|}{64.71} & 1.77 & 67.38 \\
Shufflenet\_v2\_x0\_5 & \multicolumn{1}{c|}{62.21} & 10.92 & 67.09 & \multicolumn{1}{c|}{62.67} & 4.24 & 64.76 \\
Squeezenet1\_1 & \multicolumn{1}{c|}{8.75} & 0.00 & 8.33 & \multicolumn{1}{c|}{16.13} & 1.70 & 31.86 \\ \hline
%mnasnet0\_5 & \multicolumn{1}{c|}{9.92} & 4.77 & 19.50 & \multicolumn{1}{c|}{10.42} & 0.20 & 9.24 \\
\end{tabular}
\vskip -0.1in
\end{table*}

In the previous section, we primarily used MobileNetv3-small, which is widely used in mobile execution environment~\cite{howard2019searching,9389905}. Here, we conduct additional experiments with ShuffleNet~\cite{ma2018shufflenet} and SqueezeNet~\cite{iandola2016squeezenet}, which are also mobile friendly light-weight models. Table~\ref{table:accuracyofarch} shows the worst-case accuracy for DG and the variance and average accuracy for fairness with various models. As shown in Table~\ref{table:accuracyofarch}, HeteroSwitch always shows better worst-case accuracy compared to FedAvg, demonstrating its robustness to domain generalization problem. Although ShuffleNet exhibits a decline in average accuracy with HeteroSwitch, it shows improvement in variance. Conversely, SqueezeNet, which initially fails to learn with FedAvg, showing performance equivalent to random guessing, experiences a dramatic increase in accuracy with HeteroSwitch. These results suggest that, with some adjustments, Shufflenet could work even better with our method if the observed effects are not inherent to the model structure. In short, HeteroSwitch can be used to a variety of mobile-friendly models dealing with the problems caused by system-induced data heterogeneity.

\subsection{Impact on Realistic FL Dataset}
\label{sec:withFlair}

\begin{table}[]
\caption{Evaluation of HeteroSwitch with Flair}
\label{table:flair}
\vskip 0.15in
\centering
\fontsize{9pt}{10pt}\selectfont
\begin{tabular}{|l|c|c|}
\hline
\textbf{Method} & \textbf{Averaged Precision (\%)} & \textbf{Variance} \\ \hline
(Baseline) FedAvg& 47.72 & 265.79 \\
HeteroSwitch & \textbf{47.80} & \textbf{249.02} \\
q-FedAvg & 47.25 & 257.32 \\
FedProx & 47.07 & 313.67 \\ \hline
\end{tabular}
\vskip -0.1in
\end{table}

We evaluate HeteroSwitch with a realistic FL dataset, Flair~\cite{song2022flair}. The distribution of averaged precision (AP) across device types is shown in Table~\ref{table:flair}. Because of the increased complexity of system-induced data heterogeneity, FedAvg exhibits a diverse AP distribution across device types.

As the problem becomes more complex, q-FedAvg and FedProx fail to handle the high variance properly. q-FedAvg reduces variance by 3.2\% at the expense of averaged precision compared to the baseline. FedProx performs worse with a 1.4\% decline in averaged precision and an 18\% increase in variance compared to the baseline, indicating its insufficiency to address the problem. In contrast, HeteroSwitch reduces the variance by 6.3\% improving the averaged precision by 0.2\%. This indicates that HeteroSwitch still mitigates system-induced data heterogeneity even in the presence of more diverse device types in the wild.

\subsection{Impact on Synthetic Dataset}
\label{sec:with Synthetic}
\begin{figure}[t]
  \centering
  \includegraphics[width=0.95\columnwidth]{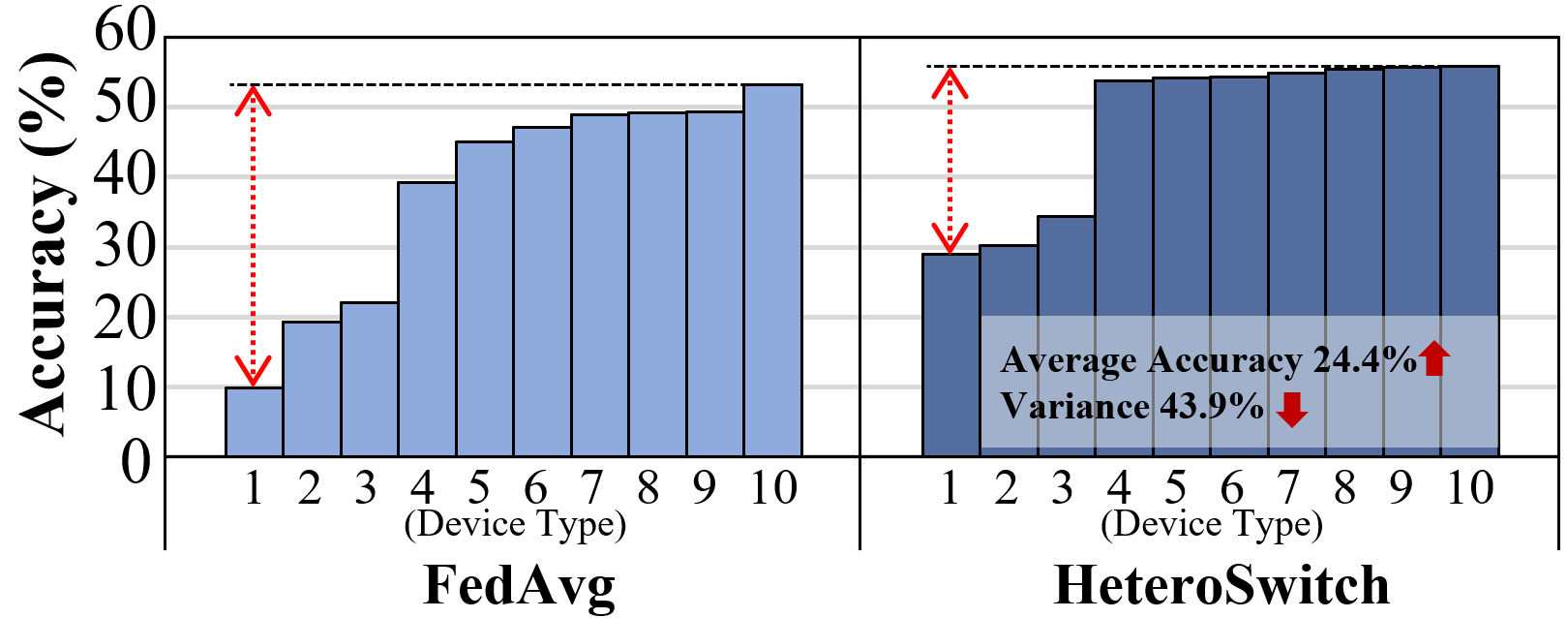}
  \caption{Model accuracy on 10 different synthetically generated device types with CIFAR-100}
  \label{fig:Cifar}
\end{figure}

In alignment with our findings on collected dataset, we further explore the impact of system-induced heterogeneity using the synthetic CIFAR-100 dataset~\cite{krizhevsky2009learning}. To emulate the diverse characteristics observed in real device data (Section~\ref{sec:breakdown2}), we inject the heterogeneity into the CIFAR-100 dataset by implementing 10 different randomized settings for contrast, brightness, saturation, and hue. A simple CNN model is used to quantify the impact of these modifications in FL setting. The distribution of accuracy across synthetic device types with FedAvg and HeteroSwitch is illustrated in Figure~\ref{fig:Cifar}.

Under synthetic condition, FedAvg achieves an average accuracy of 38.34\% but exhibits a variance of 212.97 across the synthetic devices. This scenario underscores the challenges posed by system-induced data heterogeneity, as certain device types show decreased accuracy. In contrast, HeteroSwitch significantly outperforms FedAvg, enhancing accuracy by 24.4\% and reducing variance by 43.9\%, demonstrating its effectiveness in managing device heterogeneity.

\subsection{Impact on Non-vision Dataset}
\label{sec:withECG}
We expand our evaluation to non-vision data, an Electrocardiogram (ECG) dataset. This dataset comprises data from four distinct sensor types, each introducing unique noise patterns and thereby contributing to heterogeneity in the dataset~\cite{vollmer2022simultaneous}. A simple DNN model, designed to calculate heart rates from ECG signals, is used to quantify the impact of sensor heterogeneity.

Using FedAvg, heart rate predictions from the same individual ECG data show a significant divergence, with an average deviation of 31.8\% due to sensor variability. In contrast, HeteroSwitch, equipped with a Random-Gaussian Filter, notably reduces this deviation to 18.3\%. This demonstrates that system-induced data heterogeneity is not confined to vision data and highlights the capability of HeteroSwitch to mitigate such heterogeneity across diverse data types.
\section{Related Work}

\textbf{System Heterogeneity in ML:}
In Centralized Training, several research have shown that neural networks can be vulnerable to system heterogeneity. --- e.g., they may suffer quality distortions like blurring and noise~\cite{dodge2016understanding}, be affected by simulated camera parameters such as pixel size and exposure~\cite{liu2020neural,} or display unstable inference with different devices if trained with ImageNet~\cite{cidon2021characterizing}. However, no previous work has thoroughly explored which specific steps within the signal processing process introduce biases into the ML training using real devices. Moreover, we are the first to highlight the potential challenges this could pose within a FL context, where a variety of devices can participate.

\textbf{Data Heterogeneity in FL:}
Federated Learning, with an increased number of clients, inherently produces diverse data, reflecting variations in class distribution, participant geography, culture, and device usage patterns~\cite{kairouz2021advances}, leading to data heterogeneity. Prior works have tried to mitigate this user-induced heterogeneity by adopting a regularization method to local models~\cite{li2020federated}, sharing a small amount of public data across local clients~\cite{zhao2018federated, mansour2020three}, or employing weighted averaging for model aggregation~\cite{li2019fair}. Despite these efforts, 
differences between devices remain evident even after addressing the user-induced heterogeneity. Given that different causes of data heterogeneity result in unique characteristics in the feature space (where each potentially requires own distinct solution) we focus on the system-induced heterogeneity in this work. To the best of our knowledge, our work is the first to analyze the problems caused by system-induced data heterogeneity in FL where a range of device types can participate in the learning process and propose a solution that addresses the aforementioned challenges while still preserving the data privacy in FL.

\section{Conclusion}
In this paper, we first introduce \textit{system-induced data heterogeneity} in FL. By using the dataset that we created, we show that system-induced data heterogeneity can negatively affect the accuracy of FL models. We also demonstrate the fairness and domain generalization problems which can stem from system-induced data heterogeneity in realistic execution scenarios of FL. To mitigate system-induced data heterogeneity, we propose HeteroSwitch, a selective generalization technique based on the ISP transformation and SWAD. Our evaluation results demonstrate that HeteroSwitch reduces the variance of accuracy by 79.5\% and improves the worst out-of-distribution (OOD) accuracy by 5.8\%, compared to the baseline. We believe our work will pave the path forward by enabling future work on system-induced data heterogeneity in a variety of realistic FL execution environment for a practical deployment of FL.

\section*{Acknowledgements}

This work was supported in part by National Research Foundation of Korea (NRF) grants funded by the Korea government (MSIT) (2021R1C1C1008617 and RS-2023-00212711), ICT Creative Consilience Program through the Institute of Information \& Communications Technology Planning \& Evaluation (IITP) grant funded by the Korea government (MSIT) (IITP-2024-2020-0-01819), and ITRC (Information Technology Research Center) support program through the Institute of Information \& Communications Technology Planning \& Evaluation (IITP) grant funded by the Korea government (MSIT) (IITP-2024-RS-2023-00260091).

\nocite{langley00}

\bibliography{example_paper}

\begin{thebibliography}{60}
\providecommand{\natexlab}[1]{#1}
\providecommand{\url}[1]{\texttt{#1}}
\expandafter\ifx\csname urlstyle\endcsname\relax
  \providecommand{\doi}[1]{doi: #1}\else
  \providecommand{\doi}{doi: \begingroup \urlstyle{rm}\Url}\fi

\bibitem[{Arena Com, Ltd.}(2022)]{gsmarena}
{Arena Com, Ltd.}
\newblock Gsmarena.
\newblock \url{https://www.gsmarena.com/}, 2022.
\newblock Accessed: 2023-07-11.

\bibitem[Brisimi et~al.(2018)Brisimi, Chen, Mela, Olshevsky, Paschalidis, and Shi]{brisimi2018federated}
Brisimi, T.~S., Chen, R., Mela, T., Olshevsky, A., Paschalidis, I.~C., and Shi, W.
\newblock Federated learning of predictive models from federated electronic health records.
\newblock \emph{International journal of medical informatics}, 112:\penalty0 59--67, 2018.

\bibitem[Buckler et~al.(2017)Buckler, Jayasuriya, and Sampson]{buckler2017reconfiguring}
Buckler, M., Jayasuriya, S., and Sampson, A.
\newblock Reconfiguring the imaging pipeline for computer vision.
\newblock In \emph{Proceedings of the IEEE International Conference on Computer Vision}, pp.\  975--984, 2017.

\bibitem[Cannistra()]{binning}
Cannistra, S.
\newblock Pixel binning.
\newblock \url{http://www.starrywonders.com/binning.html}.
\newblock Accessed: 2023-08-06.

\bibitem[Cha et~al.(2021)Cha, Chun, Lee, Cho, Park, Lee, and Park]{cha2021swad}
Cha, J., Chun, S., Lee, K., Cho, H.-C., Park, S., Lee, Y., and Park, S.
\newblock Swad: Domain generalization by seeking flat minima.
\newblock \emph{Advances in Neural Information Processing Systems}, 34:\penalty0 22405--22418, 2021.

\bibitem[Chipman et~al.(1997)Chipman, Kolaczyk, and McCulloch]{chipman1997adaptive}
Chipman, H.~A., Kolaczyk, E.~D., and McCulloch, R.~E.
\newblock Adaptive bayesian wavelet shrinkage.
\newblock \emph{Journal of the American Statistical Association}, 92\penalty0 (440):\penalty0 1413--1421, 1997.

\bibitem[Cidon et~al.(2021)Cidon, Pergament, Asgar, Cidon, and Katti]{cidon2021characterizing}
Cidon, E., Pergament, E., Asgar, Z., Cidon, A., and Katti, S.
\newblock Characterizing and taming model instability across edge devices.
\newblock \emph{Proceedings of Machine Learning and Systems}, 3:\penalty0 624--636, 2021.

\bibitem[Cohen et~al.(2017)Cohen, Afshar, Tapson, and Van~Schaik]{cohen2017emnist}
Cohen, G., Afshar, S., Tapson, J., and Van~Schaik, A.
\newblock Emnist: Extending mnist to handwritten letters.
\newblock In \emph{2017 international joint conference on neural networks (IJCNN)}, pp.\  2921--2926. IEEE, 2017.

\bibitem[Creager et~al.(2021)Creager, Jacobsen, and Zemel]{creager2021environment}
Creager, E., Jacobsen, J.-H., and Zemel, R.
\newblock Environment inference for invariant learning.
\newblock In \emph{International Conference on Machine Learning}, pp.\  2189--2200. PMLR, 2021.

\bibitem[Dekking et~al.(2005)Dekking, Kraaikamp, Lopuha{\"a}, and Meester]{dekking2005modern}
Dekking, F.~M., Kraaikamp, C., Lopuha{\"a}, H.~P., and Meester, L.~E.
\newblock \emph{A Modern Introduction to Probability and Statistics: Understanding why and how}, volume 488.
\newblock Springer, 2005.

\bibitem[Deng et~al.(2009)Deng, Dong, Socher, Li, Li, and Fei-Fei]{deng2009imagenet}
Deng, J., Dong, W., Socher, R., Li, L.-J., Li, K., and Fei-Fei, L.
\newblock Imagenet: A large-scale hierarchical image database.
\newblock In \emph{2009 IEEE conference on computer vision and pattern recognition}, pp.\  248--255. Ieee, 2009.

\bibitem[Dodge \& Karam(2016)Dodge and Karam]{dodge2016understanding}
Dodge, S. and Karam, L.
\newblock Understanding how image quality affects deep neural networks.
\newblock In \emph{2016 eighth international conference on quality of multimedia experience (QoMEX)}, pp.\  1--6. IEEE, 2016.

\bibitem[Dou et~al.(2019)Dou, Coelho~de Castro, Kamnitsas, and Glocker]{dou2019domain}
Dou, Q., Coelho~de Castro, D., Kamnitsas, K., and Glocker, B.
\newblock Domain generalization via model-agnostic learning of semantic features.
\newblock \emph{Advances in neural information processing systems}, 32, 2019.

\bibitem[Ebner(2007)]{ebner2007color}
Ebner, M.
\newblock \emph{Color constancy}, volume~7.
\newblock John Wiley \& Sons, 2007.

\bibitem[Gozdz(2010)]{FBDD}
Gozdz, J.
\newblock Fbdd denoising.
\newblock \url{https://valelab4.ucsf.edu/svn/" # "micromanager2/trunk/DeviceAdapters/" # "TetheredCam/LibRaw/internal/dcb_demosaicing.c}, 2010.
\newblock Accessed: 2023-08-06.

\bibitem[Guliani et~al.(2021)Guliani, Beaufays, and Motta]{guliani2021training}
Guliani, D., Beaufays, F., and Motta, G.
\newblock Training speech recognition models with federated learning: A quality/cost framework.
\newblock In \emph{ICASSP 2021-2021 IEEE International Conference on Acoustics, Speech and Signal Processing (ICASSP)}, pp.\  3080--3084. IEEE, 2021.

\bibitem[Hansen et~al.(2021)Hansen, Vilkin, Krustalev, Imber, Talagala, Hanwell, Mattina, and Whatmough]{hansen2021isp4ml}
Hansen, P., Vilkin, A., Krustalev, Y., Imber, J., Talagala, D., Hanwell, D., Mattina, M., and Whatmough, P.~N.
\newblock Isp4ml: The role of image signal processing in efficient deep learning vision systems.
\newblock In \emph{2020 25th International Conference on Pattern Recognition (ICPR)}, pp.\  2438--2445. IEEE, 2021.

\bibitem[Hejazinia et~al.(2022)Hejazinia, Huba, Leontiadis, Maeng, Malek, Melis, Mironov, Nasr, Wang, and Wu]{hejazinia2022fel}
Hejazinia, M., Huba, D., Leontiadis, I., Maeng, K., Malek, M., Melis, L., Mironov, I., Nasr, M., Wang, K., and Wu, C.-J.
\newblock Fel: High capacity learning for recommendation and ranking via federated ensemble learning.
\newblock \emph{arXiv preprint arXiv:2206.03852}, 2022.

\bibitem[Hirakawa \& Parks(2005)Hirakawa and Parks]{hirakawa2005adaptive}
Hirakawa, K. and Parks, T.~W.
\newblock Adaptive homogeneity-directed demosaicing algorithm.
\newblock \emph{Ieee transactions on image processing}, 14\penalty0 (3):\penalty0 360--369, 2005.

\bibitem[Howard et~al.(2019)Howard, Sandler, Chu, Chen, Chen, Tan, Wang, Zhu, Pang, Vasudevan, et~al.]{howard2019searching}
Howard, A., Sandler, M., Chu, G., Chen, L.-C., Chen, B., Tan, M., Wang, W., Zhu, Y., Pang, R., Vasudevan, V., et~al.
\newblock Searching for mobilenetv3.
\newblock In \emph{Proceedings of the IEEE/CVF international conference on computer vision}, pp.\  1314--1324, 2019.

\bibitem[Huba et~al.(2022)Huba, Nguyen, Malik, Zhu, Rabbat, Yousefpour, Wu, Zhan, Ustinov, Srinivas, et~al.]{huba2022papaya}
Huba, D., Nguyen, J., Malik, K., Zhu, R., Rabbat, M., Yousefpour, A., Wu, C.-J., Zhan, H., Ustinov, P., Srinivas, H., et~al.
\newblock Papaya: Practical, private, and scalable federated learning.
\newblock \emph{Proceedings of Machine Learning and Systems}, 4:\penalty0 814--832, 2022.

\bibitem[Iandola et~al.(2016)Iandola, Han, Moskewicz, Ashraf, Dally, and Keutzer]{iandola2016squeezenet}
Iandola, F.~N., Han, S., Moskewicz, M.~W., Ashraf, K., Dally, W.~J., and Keutzer, K.
\newblock Squeezenet: Alexnet-level accuracy with 50x fewer parameters and< 0.5 mb model size.
\newblock \emph{arXiv preprint arXiv:1602.07360}, 2016.

\bibitem[Izmailov et~al.(2018)Izmailov, Podoprikhin, Garipov, Vetrov, and Wilson]{izmailov2018averaging}
Izmailov, P., Podoprikhin, D., Garipov, T., Vetrov, D., and Wilson, A.~G.
\newblock Averaging weights leads to wider optima and better generalization.
\newblock \emph{arXiv preprint arXiv:1803.05407}, 2018.

\bibitem[Kairouz et~al.(2021)Kairouz, McMahan, Avent, Bellet, Bennis, Bhagoji, Bonawitz, Charles, Cormode, Cummings, et~al.]{kairouz2021advances}
Kairouz, P., McMahan, H.~B., Avent, B., Bellet, A., Bennis, M., Bhagoji, A.~N., Bonawitz, K., Charles, Z., Cormode, G., Cummings, R., et~al.
\newblock Advances and open problems in federated learning.
\newblock \emph{Foundations and Trends{\textregistered} in Machine Learning}, 14\penalty0 (1--2):\penalty0 1--210, 2021.

\bibitem[Karimireddy et~al.(2020)Karimireddy, Kale, Mohri, Reddi, Stich, and Suresh]{karimireddy2020scaffold}
Karimireddy, S.~P., Kale, S., Mohri, M., Reddi, S., Stich, S., and Suresh, A.~T.
\newblock Scaffold: Stochastic controlled averaging for federated learning.
\newblock In \emph{International Conference on Machine Learning}, pp.\  5132--5143. PMLR, 2020.

\bibitem[Kim \& Wu(2020)Kim and Wu]{kim2020autoscale}
Kim, Y.~G. and Wu, C.-J.
\newblock Autoscale: Energy efficiency optimization for stochastic edge inference using reinforcement learning.
\newblock In \emph{2020 53rd Annual IEEE/ACM International Symposium on Microarchitecture (MICRO)}, pp.\  1082--1096. IEEE, 2020.

\bibitem[Kim \& Wu(2021)Kim and Wu]{kim2021autofl}
Kim, Y.~G. and Wu, C.-J.
\newblock Autofl: Enabling heterogeneity-aware energy efficient federated learning.
\newblock In \emph{MICRO-54: 54th Annual IEEE/ACM International Symposium on Microarchitecture}, pp.\  183--198, 2021.

\bibitem[Krizhevsky et~al.(2009)Krizhevsky, Hinton, et~al.]{krizhevsky2009learning}
Krizhevsky, A., Hinton, G., et~al.
\newblock Learning multiple layers of features from tiny images.
\newblock 2009.

\bibitem[Krueger et~al.(2021)Krueger, Caballero, Jacobsen, Zhang, Binas, Zhang, Le~Priol, and Courville]{krueger2021out}
Krueger, D., Caballero, E., Jacobsen, J.-H., Zhang, A., Binas, J., Zhang, D., Le~Priol, R., and Courville, A.
\newblock Out-of-distribution generalization via risk extrapolation (rex).
\newblock In \emph{International Conference on Machine Learning}, pp.\  5815--5826. PMLR, 2021.

\bibitem[Langley(2000)]{langley00}
Langley, P.
\newblock Crafting papers on machine learning.
\newblock In Langley, P. (ed.), \emph{Proceedings of the 17th International Conference on Machine Learning (ICML 2000)}, pp.\  1207--1216, Stanford, CA, 2000. Morgan Kaufmann.

\bibitem[LeCun(1998)]{lecun1998mnist}
LeCun, Y.
\newblock The mnist database of handwritten digits.
\newblock \emph{http://yann. lecun. com/exdb/mnist/}, 1998.

\bibitem[Li et~al.(2019)Li, Sanjabi, Beirami, and Smith]{li2019fair}
Li, T., Sanjabi, M., Beirami, A., and Smith, V.
\newblock Fair resource allocation in federated learning.
\newblock \emph{arXiv preprint arXiv:1905.10497}, 2019.

\bibitem[Li et~al.(2020)Li, Sahu, Zaheer, Sanjabi, Talwalkar, and Smith]{li2020federated}
Li, T., Sahu, A.~K., Zaheer, M., Sanjabi, M., Talwalkar, A., and Smith, V.
\newblock Federated optimization in heterogeneous networks.
\newblock \emph{Proceedings of Machine learning and systems}, 2:\penalty0 429--450, 2020.

\bibitem[Li et~al.(2022)Li, Xu, Cao, Liu, Zhang, Chen, and Dai]{li2022integrated}
Li, Z., Xu, X., Cao, X., Liu, W., Zhang, Y., Chen, D., and Dai, H.
\newblock Integrated cnn and federated learning for covid-19 detection on chest x-ray images.
\newblock \emph{IEEE/ACM Transactions on Computational Biology and Bioinformatics}, 2022.

\bibitem[Lin(2003)]{lin2003pixel}
Lin, C.-k.
\newblock Pixel grouping for color filter array demosaicing, 2003.

\bibitem[Liu et~al.(2020)Liu, Lian, Farrell, and Wandell]{liu2020neural}
Liu, Z., Lian, T., Farrell, J., and Wandell, B.~A.
\newblock Neural network generalization: The impact of camera parameters.
\newblock \emph{IEEE Access}, 8:\penalty0 10443--10454, 2020.

\bibitem[Ma et~al.(2018)Ma, Zhang, Zheng, and Sun]{ma2018shufflenet}
Ma, N., Zhang, X., Zheng, H.-T., and Sun, J.
\newblock Shufflenet v2: Practical guidelines for efficient cnn architecture design.
\newblock In \emph{Proceedings of the European conference on computer vision (ECCV)}, pp.\  116--131, 2018.

\bibitem[Maeng et~al.(2022)Maeng, Lu, Melis, Nguyen, Rabbat, and Wu]{maeng2022towards}
Maeng, K., Lu, H., Melis, L., Nguyen, J., Rabbat, M., and Wu, C.-J.
\newblock Towards fair federated recommendation learning: Characterizing the inter-dependence of system and data heterogeneity.
\newblock In \emph{Proceedings of the 16th ACM Conference on Recommender Systems}, pp.\  156--167, 2022.

\bibitem[Mansour et~al.(2020)Mansour, Mohri, Ro, and Suresh]{mansour2020three}
Mansour, Y., Mohri, M., Ro, J., and Suresh, A.~T.
\newblock Three approaches for personalization with applications to federated learning.
\newblock \emph{arXiv preprint arXiv:2002.10619}, 2020.

\bibitem[McMahan et~al.(2017)McMahan, Moore, Ramage, Hampson, and y~Arcas]{mcmahan2017communication}
McMahan, B., Moore, E., Ramage, D., Hampson, S., and y~Arcas, B.~A.
\newblock Communication-efficient learning of deep networks from decentralized data.
\newblock In \emph{Artificial intelligence and statistics}, pp.\  1273--1282. PMLR, 2017.

\bibitem[Mohri et~al.(2019)Mohri, Sivek, and Suresh]{mohri2019agnostic}
Mohri, M., Sivek, G., and Suresh, A.~T.
\newblock Agnostic federated learning.
\newblock In \emph{International Conference on Machine Learning}, pp.\  4615--4625. PMLR, 2019.

\bibitem[Morovi{\v{c}}(2008)]{morovivc2008color}
Morovi{\v{c}}, J.
\newblock \emph{Color gamut mapping}.
\newblock John Wiley \& Sons, 2008.

\bibitem[Pessach \& Shmueli(2022)Pessach and Shmueli]{pessach2022review}
Pessach, D. and Shmueli, E.
\newblock A review on fairness in machine learning.
\newblock \emph{ACM Computing Surveys (CSUR)}, 55\penalty0 (3):\penalty0 1--44, 2022.

\bibitem[Qian et~al.(2021)Qian, Ning, and Hu]{9389905}
Qian, S., Ning, C., and Hu, Y.
\newblock Mobilenetv3 for image classification.
\newblock In \emph{2021 IEEE 2nd International Conference on Big Data, Artificial Intelligence and Internet of Things Engineering (ICBAIE)}, pp.\  490--497, 2021.
\newblock \doi{10.1109/ICBAIE52039.2021.9389905}.

\bibitem[Sagawa et~al.(2019)Sagawa, Koh, Hashimoto, and Liang]{sagawa2019distributionally}
Sagawa, S., Koh, P.~W., Hashimoto, T.~B., and Liang, P.
\newblock Distributionally robust neural networks for group shifts: On the importance of regularization for worst-case generalization.
\newblock \emph{arXiv preprint arXiv:1911.08731}, 2019.

\bibitem[Scheuerman et~al.(2021)Scheuerman, Hanna, and Denton]{scheuerman2021datasets}
Scheuerman, M.~K., Hanna, A., and Denton, E.
\newblock Do datasets have politics? disciplinary values in computer vision dataset development.
\newblock \emph{Proceedings of the ACM on Human-Computer Interaction}, 5\penalty0 (CSCW2):\penalty0 1--37, 2021.

\bibitem[Segu et~al.(2023)Segu, Tonioni, and Tombari]{segu2023batch}
Segu, M., Tonioni, A., and Tombari, F.
\newblock Batch normalization embeddings for deep domain generalization.
\newblock \emph{Pattern Recognition}, 135:\penalty0 109115, 2023.

\bibitem[Shi et~al.(2023)Shi, Yu, and Leung]{shi2023towards}
Shi, Y., Yu, H., and Leung, C.
\newblock Towards fairness-aware federated learning.
\newblock \emph{IEEE Transactions on Neural Networks and Learning Systems}, 2023.

\bibitem[Shorten \& Khoshgoftaar(2019)Shorten and Khoshgoftaar]{shorten2019survey}
Shorten, C. and Khoshgoftaar, T.~M.
\newblock A survey on image data augmentation for deep learning.
\newblock \emph{Journal of big data}, 6\penalty0 (1):\penalty0 1--48, 2019.

\bibitem[Song et~al.(2022)Song, Granqvist, and Talwar]{song2022flair}
Song, C., Granqvist, F., and Talwar, K.
\newblock Flair: Federated learning annotated image repository.
\newblock \emph{Advances in Neural Information Processing Systems}, 35:\penalty0 37792--37805, 2022.

\bibitem[StatCounter(2022)]{statcounter}
StatCounter.
\newblock Mobile vendor market share.
\newblock \url{https://gs.statcounter.com/vendor-market-share/mobile/united-states-of-america/2021}, 2022.
\newblock Accessed: 2023-07-11.

\bibitem[Stokes(1996)]{stokes1996standard}
Stokes, M.
\newblock A standard default color space for the internet-srgb.
\newblock \emph{http://www. w3. org/Graphics/Color/sRGB. html}, 1996.

\bibitem[Tommasi et~al.(2017)Tommasi, Patricia, Caputo, and Tuytelaars]{tommasi2017deeper}
Tommasi, T., Patricia, N., Caputo, B., and Tuytelaars, T.
\newblock A deeper look at dataset bias.
\newblock \emph{Domain adaptation in computer vision applications}, pp.\  37--55, 2017.

\bibitem[Vollmer et~al.(2022)Vollmer, Bl{\"a}sing, Reiser, Nisser, and Buder]{vollmer2022simultaneous}
Vollmer, M., Bl{\"a}sing, D., Reiser, J., Nisser, M., and Buder, A.
\newblock Simultaneous physiological measurements with five devices at different cognitive and physical loads.
\newblock \emph{Physionet}, 101\penalty0 (23):\penalty0 215--220, 2022.

\bibitem[Wang et~al.(2022)Wang, Lan, Liu, Ouyang, Qin, Lu, Chen, Zeng, and Yu]{wang2022generalizing}
Wang, J., Lan, C., Liu, C., Ouyang, Y., Qin, T., Lu, W., Chen, Y., Zeng, W., and Yu, P.
\newblock Generalizing to unseen domains: A survey on domain generalization.
\newblock \emph{IEEE Transactions on Knowledge and Data Engineering}, 2022.

\bibitem[Wu et~al.(2019)Wu, Brooks, Chen, Chen, Choudhury, Dukhan, Hazelwood, Isaac, Jia, Jia, et~al.]{wu2019machine}
Wu, C.-J., Brooks, D., Chen, K., Chen, D., Choudhury, S., Dukhan, M., Hazelwood, K., Isaac, E., Jia, Y., Jia, B., et~al.
\newblock Machine learning at facebook: Understanding inference at the edge.
\newblock In \emph{2019 IEEE international symposium on high performance computer architecture (HPCA)}, pp.\  331--344. IEEE, 2019.

\bibitem[Wu \& Wang(2022)Wu and Wang]{wu2022node}
Wu, H. and Wang, P.
\newblock Node selection toward faster convergence for federated learning on non-iid data.
\newblock \emph{IEEE Transactions on Network Science and Engineering}, 9\penalty0 (5):\penalty0 3099--3111, 2022.

\bibitem[Wyszecki \& Stiles(2000)Wyszecki and Stiles]{wyszecki2000color}
Wyszecki, G. and Stiles, W.~S.
\newblock \emph{Color science: concepts and methods, quantitative data and formulae}, volume~40.
\newblock John wiley \& sons, 2000.

\bibitem[Zhang et~al.(2016)Zhang, Li, Qiao, Wang, Yan, Li, and Hu]{zhang2016image}
Zhang, H., Li, L., Qiao, K., Wang, L., Yan, B., Li, L., and Hu, G.
\newblock Image prediction for limited-angle tomography via deep learning with convolutional neural network.
\newblock \emph{arXiv preprint arXiv:1607.08707}, 2016.

\bibitem[Zhao et~al.(2018)Zhao, Li, Lai, Suda, Civin, and Chandra]{zhao2018federated}
Zhao, Y., Li, M., Lai, L., Suda, N., Civin, D., and Chandra, V.
\newblock Federated learning with non-iid data.
\newblock \emph{arXiv preprint arXiv:1806.00582}, 2018.

\end{thebibliography}
\bibliographystyle{mlsys2024}

%%%%%%%%%%%%%%%%%%%%%%%%%%%%%%%%%%%%%%%%%%%%%%%%%%%%%%%%%%%%%%%%%%%%%%%%%%%%%%%
%%%%%%%%%%%%%%%%%%%%%%%%%%%%%%%%%%%%%%%%%%%%%%%%%%%%%%%%%%%%%%%%%%%%%%%%%%%%%%%
% SUPPLEMENTAL CONTENT AS APPENDIX AFTER REFERENCES
%%%%%%%%%%%%%%%%%%%%%%%%%%%%%%%%%%%%%%%%%%%%%%%%%%%%%%%%%%%%%%%%%%%%%%%%%%%%%%%
%%%%%%%%%%%%%%%%%%%%%%%%%%%%%%%%%%%%%%%%%%%%%%%%%%%%%%%%%%%%%%%%%%%%%%%%%%%%%%%
\appendix
% \section{Please add supplemental material as appendix here}
\onecolumn
\section{Experimental Details}

\subsection{Hardware \& Software}
We implement FL with PyTorch and TensorFlow, and emulate it with our dataset and Flair, respectively, on our servers. \textbf{The following is the experimental environment used for our dataset 
 (Section~\ref{sec:datasetcreation}). }

\begin{compactitem}
    \item GPU/CPU models: No GPU, Intel(R) Xeon(R) Silver 4210R CPU @ 2.40GHz
    \item Amount of memory: 128GiB
    \item Operating system: Ubuntu 20.04.5
    \item Version of PyTorch: 1.12.0
\end{compactitem}

\textbf{The following is the experimental environment used for the Flair~\cite{song2022flair}.}

\begin{compactitem}
    \item GPU/CPU models: NVIDIA Tesla V100 GPU, Intel(R) Xeon(R) Gold 5218 CPU @ 2.30GHz
    \item Amount of memory: 128GiB
    \item Operating system: Ubuntu 20.04.3
    \item Version of TensorFlow: 2.9.0
    \item Version of TensorFlow-Federated: 0.20.0
\end{compactitem}

\subsection{FL Parameters}
\begin{figure}[h]
\centering
\includegraphics[width=0.9\textwidth]{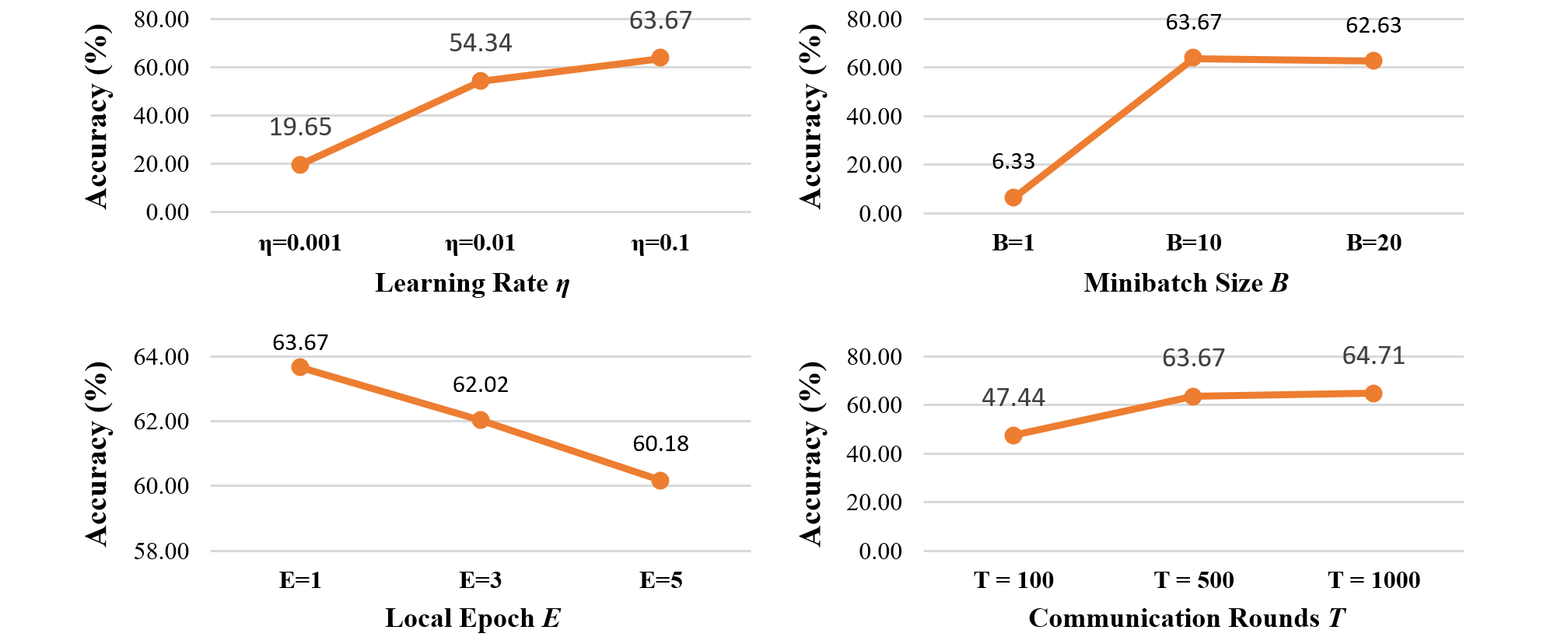}
\caption{Selection of Learning Rate, Minibatch Size, Local Epoch, and Communication Rounds} 
\label{fig:appendix1}
\end{figure}

We set the hyperparameters of FL based on the sensitivity analysis.\textbf{ The hyperparameters that we test are summarized as follows:}
\begin{compactitem}
    \item learning rate $\eta \in \{0.001, 0.01, 0.1\}$
    \item Minibatch size $B \in \{1,10,20\}$
    \item Local epochs $E \in \{1,3,5\}$
    \item Number of communication rounds $T \in \{100, 500, 1000\}$
\end{compactitem}
Fig.~\ref{fig:appendix1} shows the sensitivity analysis result on the learning rate, minibatch size, local epoch, and number communication rounds. Based on results presented in Fig.~\ref{fig:appendix1}, we select 0.1, 10, 1, and 1000 for the learning rate, minibatch size, local epoch, and the number of communication rounds, respectively, for the rest of our experiments (i.e., characterization and evaluation in Section~\ref{sec:characterize} and \ref{sec:evaluation}, respectively).

For a fair comparison with q-FedAvg and FedProx, we also configure the hyperparameters of q-FedAvg and FedProx based on the sensitivity analysis. \textbf{Below are the hyperparameter grids that we searched on for previous works(Section ~\ref{sec:Compare}).}

\begin{compactitem}
    \item q-FedAvg $q \in \{1e-6, 1e-5, 1e-4, 1e-3, 1e-2, 1e-1\}$
    \item FedProx $\mu \in \{1e-5, 1e-4, 1e-3, 1e-2, 1e-1\}$
\end{compactitem}

we use \(q = 1e-6\) and \(\mu = 1e-1\), for q-FedAvg, and FedProx, respectively.

\textbf{Below are the hyperparameter grids that we searched on for HeteroSwitch}
\begin{compactitem}
    \item Random WB $degree \in \{0.001, 0.01, 0.1, 0.5, 0.9\} $    
    \item Random Gamma $degree \in \{0.1, 0.3, 0.5, 0.7, 0.9\}$
\end{compactitem}

We set the smoothing factor for Exponential Moving Average \(\alpha = 0.9\) for \(L_{EMA}\), Random WB \(degree = 0.001 \) and Random Gamma \(degree = 0.9 \). 
%Put anything that you might normally include after the references as an appendix here, {\it not in a separate supplementary file}. Upload your final camera-ready as a single pdf, including all appendices.

%%%%%%%%%%%%%%%%%%%%%%%%%%%%%%%%%%%%%%%%%%%%%%%%%%%%%%%%%%%%%%%%%%%%%%%%%%%%%%%
%%%%%%%%%%%%%%%%%%%%%%%%%%%%%%%%%%%%%%%%%%%%%%%%%%%%%%%%%%%%%%%%%%%%%%%%%%%%%%%

\end{document}